\def\buildcombined{1}

\documentclass{article}
\usepackage[utf8]{inputenc}
\usepackage{authblk}
\usepackage{setspace}
\usepackage[margin=1.25in]{geometry}
\usepackage{graphicx}
\graphicspath{ {./figures/} }
\usepackage{subcaption}
\usepackage{amsmath}
\usepackage{booktabs}
\usepackage{tabularx}
\usepackage{makecell}
\usepackage{multirow}
\usepackage{pdflscape}
\usepackage{float}
\usepackage{flafter}
\usepackage{xurl}

\usepackage[authoryear,round,semicolon]{natbib}
\ifdefined\buildcombined\else
\usepackage{xr}
\fi
\usepackage[hidelinks]{hyperref}

\title{Annual Earth-observation embeddings encode wildfire disturbance and support simplified burned area mapping}

\author[1,2,3*]{Jovana Knezevic}
\author[4]{Clement Atzberger}
\author[3]{Zhengpeng Feng}
\author[5]{Adam F. A. Pellegrini}
\author[1,3]{Srinivasan Keshav}
\author[1,2,3*]{David Coomes}

\affil[1]{Conservation Research Institute, University of Cambridge, Cambridge, United Kingdom.}
\affil[2]{Department of Plant Sciences, University of Cambridge, Cambridge, United Kingdom.}
\affil[3]{Department of Computer Science and Technology, University of Cambridge, Cambridge, United Kingdom.}
\affil[4]{dClimate Labs, New York, US.}
\affil[5]{Department of Earth System Science, Stanford University, Stanford, CA, USA.}
\affil[*]{Address correspondence to: jk871@cam.ac.uk and dac18@cam.ac.uk}

\date{}

\begin{document}

\maketitle

\begin{abstract}
Medium-resolution (10–30 m) burned area mapping is vital for monitoring wildfires and their impacts, but remains difficult to scale. Existing methods require either curated fire-specific imagery or dense time-series analysis. Here, we tested whether annual Earth-observation embeddings retain wildfire disturbance signals sufficiently to map burned areas without either requirement. Using Tessera and AlphaEarth embeddings, we tested individual burn-scar delineation, mapping of all same-year fires within an area, regional wall-to-wall mapping, cross-continental transfer, and intra-annual fire timing. Tessera strongly encoded wildfire disturbance, allowing even linear models to separate burned from unburned pixels; the signal was weaker in AlphaEarth. Models trained on a single Tessera embedding matched or exceeded equivalent models using paired pre- and post-fire HLS imagery, and outperformed post-fire imagery alone. The same approach mapped all same-year fires within benchmark scenes (F1 = 0.90). Applied across California, with no California fire data used for downstream training, it recovered 97\% of reference burned area and detected substantially more small and medium-sized fires than GABAM or MCD64A1. Separately, a model trained on 2018--2021 US fires transferred without retraining to 88 European fires from 2024--2025 (F1 = 0.88). For well-detected fires, ignition timing was recovered with a mean absolute error of 13 days. Performance declined for fires ignited near the end of the calendar year, and wall-to-wall deployment produced systematic false positives in some unseen landscapes. Annual embeddings nevertheless achieve high segmentation accuracy while moving the burden of dense time series processing upstream, providing a promising path towards simpler regional burned area mapping.
\end{abstract}


\section{Introduction}

Accurate mapping of burned areas provides a foundation for evaluating the impacts of fires on local, regional and global landscapes \citep{mouillotTenYearsGlobal2014}. Satellite data enable burned areas to be mapped over large scales,  and these products are used to characterize changes in fire regimes \citep{andelaHumandrivenDeclineGlobal2017b}, assess post-fire vegetation recovery \citep{noleBiogeographicVariabilityWildfire2022}, quantify carbon emissions caused by fires \citep{vanderwerfGlobalFireEmissions2017}, and estimate wildfire exposure and hazard \citep{chenWildfireRiskGlobal2024,shahNewFireDanger2022, verdeAssessmentValidationWildfire2010}.  Yet current global burned area products are produced at relatively coarse spatial resolution, from 250--500 m raster products \citep{giglioNASAVIIRSBurned2025, giglioCollection6MODIS2018a} to $0.25^\circ$ gridded datasets \citep{earthsciencedatasystemsGlobalFireEmissions2025}. At these scales, fire perimeters cannot be resolved in detail \citep{humberAssessingShapeAccuracy2020, hallMODISbasedBurnedArea2016}, small fires are systematically omitted \citep{ramoAfricanBurnedArea2021a,fornaccaIncludingSmallFires2025,boschettiGlobalValidationCollection2019}, and accuracy deteriorates in fragmented landscapes \citep{larisSpatiotemporalProblemsDetecting2005, rodriguesHowWellGlobal2019, boschettiGlobalValidationCollection2019}. Small islands of unburned vegetation, which act as ecologically important refuges and seed sources for post-fire recovery, are also far below the resolution of coarse products \citep{coopContributionsFireRefugia2019}. Reflecting these limitations, a recent review identified the need to develop 10--30 m (i.e. decametric)  products \citep{radeloffNeedVisionGlobal2024a}. Decametric resolution improves burned area and fire emission estimates \citep{ramoAfricanBurnedArea2021a, qiHighResolution302024a}, captures heterogeneous burning regimes \citep{liuBurnedAreaDetection2018}, and maps burned areas in greater spatial detail \citep{royLandsat8Sentinel2Burned2019} and in closer agreement with high-resolution reference imagery \citep{stroppianaSpatialAgreementBurned2025}. Global Annual Burned Area Map (GABAM), a 30 m Landsat-derived product, remains the only global burned area map at decametric resolution \citep{long30ResolutionGlobal2019a}, and analyses based on it have shown that finer resolution substantially increases the detection of small fires missed by coarser products \citep{fornaccaIncludingSmallFires2025,qiHighResolution302024a}. Sentinel-1 and Sentinel-2 offer higher spatial and temporal resolution than Landsat and have shown considerable promise for burned area mapping in regional studies \citep{rotetaDevelopmentSentinel2Burned2019b, ramoAfricanBurnedArea2021a, verhegghenPotentialSentinelSatellites2016, gaveauRefinedBurnedareaMapping2021, deshpandeDetectingQuantifyingResidue2022}, yet no global burned area product based on these sensors currently exists.

Existing burned area mapping approaches can be viewed as addressing two related but distinct tasks: delineating known fire events and discovering all burned area across a region.  \textbf{Event-based} approaches map the burned area of individual, known fires from curated post-fire satellite images that contain the visible burn scar. Deep learning models achieve high segmentation accuracies in event-based studies, producing maps at 10~m resolution. These approaches work well with single post-fire images as training data \citep{seydiBurntNetWildfireBurned2022, huUniTemporalMultispectralImagery2021, fuBurnedAreaSegmentation2024, knoppDeepLearningApproach2020, sdrakaFLOGAMachineLearningReadyDataset2024}, and even better when pre- and post-fire image pairs are available \citep{suiBiAUNetWildfireBurnt2024,liuFasterBetterMore2026, sdrakaFLOGAMachineLearningReadyDataset2024}. This accuracy, however, rests on substantial task-specific data curation. For each fire, one must identify the fire timing, select suitable pre- and post-fire imagery and choose an adequate post-fire timing window: after the fire has been contained, but before the disturbance signal has substantially diminished. Because deep learning models are data-demanding, the best approaches spend significant effort constructing large collections of such curated fire-specific images for training \citep{liuFasterBetterMore2026, sdrakaFLOGAMachineLearningReadyDataset2024}. The same curation is required again at inference: a model cannot be applied to a new fire without a suitable post-fire image being supplied first. As a result, event-based approaches achieve excellent burn scar delineation, but are difficult to scale operationally, as the temporal localization burden is hidden inside the data curation.   \textbf{Region-based approaches} aim to detect and delineate all burned area across a landscape within a given period, without prior knowledge of individual fires' locations or occurrence times. Typically, this requires searching large volumes of repeated satellite observations across space and time, in which most observations contain no burned area at all. Sifting through entire time series is data-intensive and computationally expensive, significantly so at higher spatial resolutions, and every operational product limits this cost in some way, trading off spatial resolution, temporal resolution, and model complexity. MODIS MCD64A1 and VIIRS VNP64A1 accomplish wall-to-wall temporal change detection, sliding two adjacent pre- and post-change windows through the entire valid observation time series of each pixel, by operating at a hectometric  resolution of 500~m \citep{giglioCollection6MODIS2018a, giglioNASAVIIRSBurned2025}. FireCCI50, at 250~m, instead narrows the search using the MODIS active fire product, only examining 1~km grid cells in which a hotspot was detected \citep{chuviecoGenerationAnalysisNew2018a}. GABAM operates at a finer 30~m resolution, but pays for it elsewhere: temporal resolution drops to Landsat's 16-day revisit interval compared to MODIS's daily, and the per-pixel model is a small Random Forest run on each valid Landsat observation \citep{long30ResolutionGlobal2019a}. Even the most recent integration of deep learning into operational mapping confirms this cost structure. Liu et al, for example, propose a two-stage pipeline in which a random forest first searches Sentinel-2 observations for candidate burned areas, and only the downloaded candidate regions are passed to a large dual-temporal deep network \citep{liuFasterBetterMore2026}.

A comparison of event-based and region-based approaches exposes a central tension: accurate burned area delineation is achievable when fire location and timing are already known, but regional mapping must first discover those events by processing observations across both space and time. Thus, while a global 10~m burned area product would meet a well-documented need \citep{radeloffNeedVisionGlobal2024a}, the wall-to-wall dense temporal search required to produce one remains computationally and operationally demanding.  This paper evaluates whether the rapidly advancing field of geospatial foundation modelling provides opportunities to generate global 10-m annual-resolution burned area products, bridging the divide between current approaches.   

Geospatial foundation models (GFMs) have recently emerged as a new technology that promises to simplify and scale traditional remote sensing workflows \citep{klemmerEarthEmbeddingsAIcentric2025}. GFMs are large machine learning models, pre-trained on vast amounts of multi-modal remote sensing data through self-supervised learning, producing learned representations that can support a variety of geospatial tasks. Their main promise is plug-and-play ease of use and scalability, alleviating the burden of cloud masking, data processing, and task-specific feature construction. A subset of GFMs, including Tessera \citep{fengTESSERATemporalEmbeddings2026} and AlphaEarth \citep{brownAlphaEarthFoundationsEmbedding2025}, are temporal models: they summarize a whole year of satellite data into precomputed, multi-dimensional vectors, called embeddings. These spectral-temporal embeddings offer a different way to work with dense satellite time series: rather than explicitly searching through raw observations, the temporal reasoning is performed once, upstream, when the representation is generated. If wildfire disturbance information is adequately captured by this summarization, the dense temporal search at the heart of region-based mapping could be replaced by reading the disturbance out of a single annual representation. To what extent that information is retained remains an open question: temporal embeddings have so far been validated mostly on undisturbed landscape properties such as land cover, crop type, and vegetation structure \citep{ballGeospatialFoundationModels2026,liuCITYREPUnifiedBenchmark2026, plasBetterTogetherEvaluating2026, maHarvestingAlphaEarthBenchmarking2026}, with initial assessments of burned area mapping only recently appearing \citep{seydiDeepLearningBasedBurned2025a,silvaEvaluatingAlphaEarthTESSERA2026}. It remains unclear whether annual embeddings can support the full burned area mapping problem: delineating entire burn scars from a single annual embedding, discovering all fires across a landscape, and retrieving when they occurred. It is also unclear whether this capability depends on how the embeddings are constructed, since models differ in sensors, architectures, and training objectives.

In this study, we treated annual embeddings from two foundation models, Tessera and AlphaEarth, as inputs to lightweight downstream models, asking whether a single-year embedding can map where fires burned without being told where and when they occurred. We focused our contributions around the following questions.
\begin{enumerate}
\item Is wildfire disturbance encoded within annual embeddings, and readily extractable?
\item Can annual embeddings match the event-based setting, delineating individual burn scars without curated post-fire imagery?
\item Can embeddings support the region-based setting, mapping all burned area across a landscape, including application to an unseen region and transfer across continents (i.e.\ zero-shot modelling)?
\item Can embeddings recover sub-annual fire timing?
\end{enumerate}
Throughout, we examined where and why embedding-based approaches fail. We addressed these questions using the HLS Burn Scars benchmark across the continental United States (2018--2021), a statewide wall-to-wall deployment in California, and 88 European fires from 2024--2025. Our results show that annual embeddings can support all four tasks, with Tessera consistently outperforming AlphaEarth in settings where the two representations were directly compared. Embedding-based workflows mapped burned area directly from the annual embedding, without event-specific imagery, dense time-series processing, or explicit comparison of pre- and post-event periods. Because much of the temporal processing is shifted upstream into a reusable representation, the downstream workflow is substantially simpler, offering a path towards more scalable medium-resolution burned area mapping. 

\section{Datasets}

\subsection{Annual embeddings from Tessera and AlphaEarth}

We used annual precomputed temporal embeddings from Tessera and AlphaEarth as fixed inputs for burned area segmentation and fire timing. The embedding models were not fine-tuned; all supervised classifiers and segmentation models trained on top of embeddings are referred to as downstream models. We focused primarily on Tessera, using AlphaEarth as a complementary comparison.

Tessera is a GFM that provides publicly available annual embeddings \citep{fengTESSERATemporalEmbeddings2026}. It was trained on globally distributed time series of Sentinel-1 and Sentinel-2 data to produce embeddings that are robust to missing and cloudy observations \citep{lisaiusUsingBarlowTwins2024}. Each embedding is a 128-dimensional unnormalized vector that represents a spectro-temporal summary of one calendar year for a 10$\times$10~m pixel. We used Tessera V1 annual embeddings for 2017–2025, downloaded through the geotessera library (\url{https://github.com/ucam-eo/geotessera}).

AlphaEarth also provides precomputed annual embeddings at 10 m resolution \citep{brownAlphaEarthFoundationsEmbedding2025}. AlphaEarth was trained on Sentinel-1, Sentinel-2 and Landsat data, while also using other data sources as prediction targets, including lidar, elevation, climate, land cover and text from Wikipedia \citep{brownAlphaEarthFoundationsEmbedding2025}. It produces 64-dimensional unit-norm embeddings. Unlike pixel-level Tessera, AlphaEarth is trained on image patches, so each pixel embedding incorporates surrounding spatial context at approximately kilometre scale. The v1 embeddings were downloaded from the Taylor Geospatial Cloud-Optimized GeoTIFF mirror on Source Cooperative (\url{https://source.coop/tge-labs/aef}).

The precomputed products from both models summarize a calendar year (1 January–31 December). We denote $E_0$ for fire year embeddings, the calendar year in which the target fire ignition occurred; $E_{-1}$ for embeddings from the year preceding fire, and $E_{+1}$ for embeddings from the year following the fire.

\subsection{Continental United States dataset}
\label{sec:hls-burn-scars}

\textit{HLS Burn Scars benchmark.}
To compare annual embeddings with an event-based workflow, we used the Harmonized Landsat and Sentinel-2 (HLS) Burn Scars benchmark \citep{HLSBurnScarsDataset}. It contains 804 chips (512$\times$512-pixel image tiles covering 15$\times$15~km at 30~m resolution), each centered on a fire event and paired with a post-fire HLS image and a burned area mask derived from Monitoring Trends in Burn Severity (MTBS). Post-fire images were acquired 1--3 months after ignition, selected for \textless10\% cloud cover, and visually checked for the presence of the burn scar \citep{jakubikFoundationModelsGeneralist2023}. Depending on the size of the fire, the chips may only capture part of the event.

\vspace{\baselineskip}
\noindent\textit{Supplementing the HLS Burn Scars database with pre-fire imagery.}
To create the paired pre/post baseline, we generated pre-fire composites from HLS Sentinel-2 (S30) v2.0 surface reflectance \citep{neigh_ju_roger_skakun_vermote_claverie_dungan_yin_freitag_justice_2021, juHarmonizedLandsatSentinel22025}, obtained through Google Earth Engine (GEE). Observations were drawn from the 30 days preceding MTBS ignition, extended to 60 days when valid coverage remained below 95\%. Cloud, shadow, and snow were masked, and valid observations were combined using a temporally weighted medoid. Full compositing and missing-data procedures are provided in Section~\ref{supp:prefire-hls} of the Supplementary Material.

\vspace{\baselineskip}
\noindent\textit{MTBS reference data and label refinement.}
We used fire perimeters, severity maps, and ignition dates from MTBS to refine benchmark burned area masks and construct masks for annual-fire evaluation \citep{eidenshinkProjectMonitoringTrends2007}. MTBS maps large US fires ($>\sim$400~ha for the western US and $>\sim$200~ha in the eastern US) at 30~m resolution using Landsat imagery and expert review \citep{eidenshinkProjectMonitoringTrends2007}. We accessed two MTBS products through GEE: Burned Area Boundaries (\path{USFS/GTAC/MTBS/burned_area_boundaries/v1}) and Annual Burn Severity Mosaics (\path{USFS/GTAC/MTBS/annual_burn_severity_mosaics/v1}) \citep{usdaforestserviceMonitoringTrendsBurnb}. Ignition dates were used as event dates in timing analyses.

Because MTBS perimeters may contain unburned or spectrally ambiguous areas, we reconstructed the HLS Burn Scars labels using the corresponding MTBS severity mosaics. Classes 2–4 (low, moderate, and high severity) were treated as burned, while classes 0, 1, and 5 (unburned, unburned to low severity, and increased greenness) were treated as unburned, consistent with previous studies \citep{shiHistoricalCoverTrends2018,huLargescaleBurnSeverity2023}. Examples are shown in Figure S1. Compatible severity masks could not be constructed for 11 chips, leaving 793 chips for subsequent analyses unless otherwise stated.

Each benchmark chip was constructed around a single target fire, but could also intersect additional fires that occurred during the same calendar year. These additional fires are relevant for annual embeddings, which summarize the full calendar year and may encode any fire that occurred during that time. We therefore distinguished the original target fire from the union of all same-year fires intersecting each chip; the corresponding single-fire and annual-fire evaluation protocols are described in Section~\ref{sec:reference-labels}. When all MTBS fires from the corresponding calendar year were included, however, a single fire could intersect more than one chip. We therefore distinguish three units: a \emph{chip} is one benchmark image tile; a \emph{unique fire event} is one distinct MTBS fire; and a \emph{fire--chip pair} is one intersection between a fire event and a chip. Across the 793 chips, there were 1,162 unique MTBS events and 1,314 fire–chip pairs. Analysis-specific eligibility filters produced the smaller cohorts summarized in Table~\ref{tab:experimental-overview}.

\begin{figure}[H]
\centering
\includegraphics[width=\textwidth]{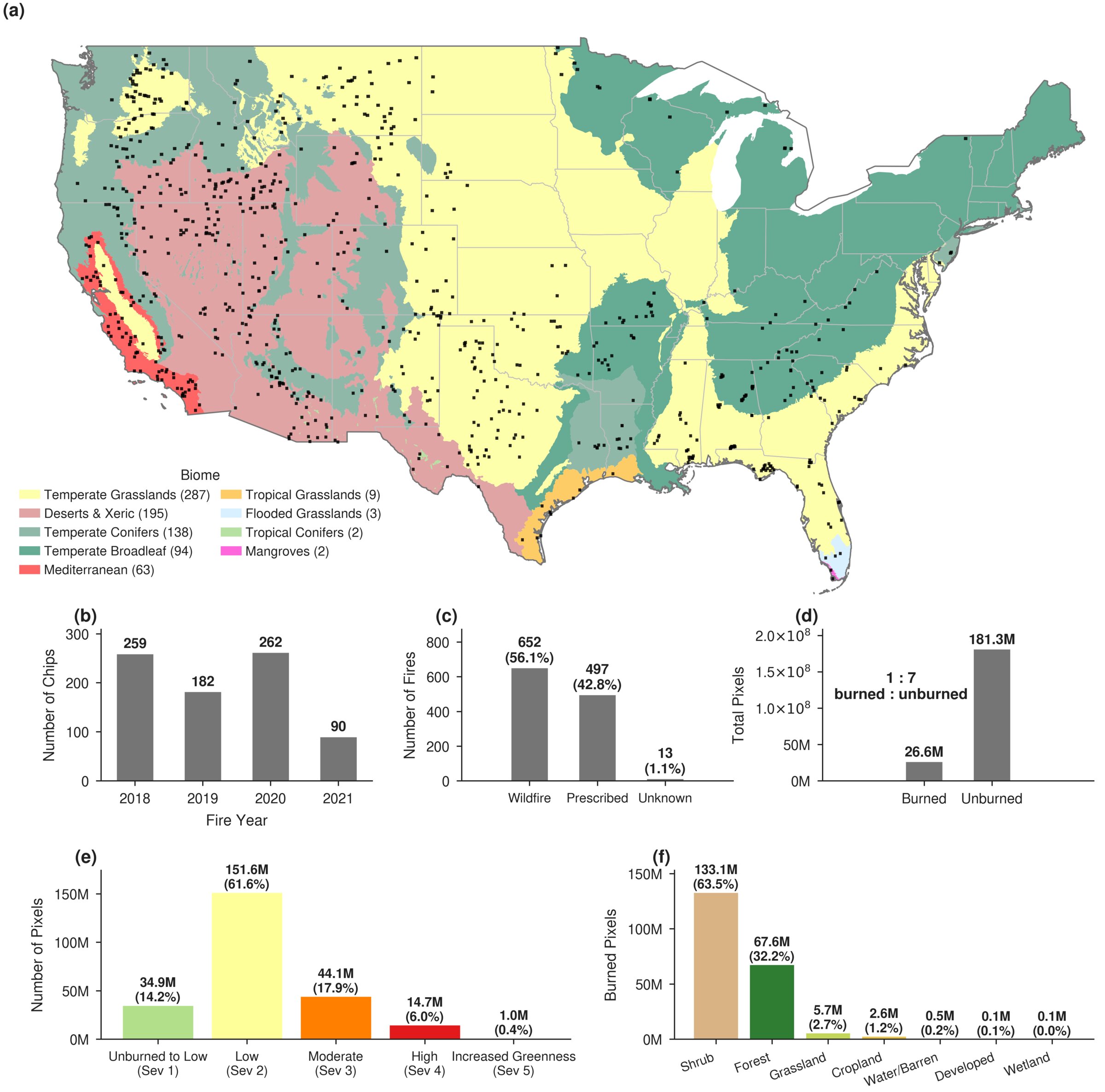}
\caption{Distribution and composition of the HLS Burn Scars benchmark across the contiguous United States (2018--2021). (a)~Chip locations by biome; (b)~chips by fire year; (c)~fires by incident type; (d)~burned and unburned pixels; (e)~MTBS burn severity; and (f)~pre-fire land cover. Map lines delineate study areas and do not necessarily depict accepted national boundaries.}
\label{fig:study_area}
\end{figure}

\vspace{\baselineskip}
\noindent\textit{Benchmark composition.}
The chips span fires across the contiguous United States from 2018–2021, with most events occurring in the western US (Figure~\ref{fig:study_area}). We obtained biome information from RESOLVE Ecoregions 2017 \citep{dinersteinEcoregionBasedApproachProtecting2017}, and pre-fire land cover from Land Change Monitoring, Assessment, and Projection (LCMAP) v1.3 \citep{brownLessonsLearnedImplementing2020a}, both via GEE. Most chips represent fires in temperate, xeric, and Mediterranean biomes. The majority of the burned area occurred in shrublands (63.5\%) and forests (32.2\%), and the cohort includes both wildfires and prescribed fires. Pixels that burned at low severity account for 61.6\% of the total burned area, while high-severity pixels represent only 6.0\%. The dataset is also strongly class-imbalanced, with an approximately 1:7 ratio of burned to unburned pixels.

\subsection{Californian dataset}
\label{sec:burned-area-products}

For wall-to-wall evaluation, we mapped burned area across California in 2021 using two complementary reference datasets: 37 fires represented by MTBS severity data and 386 wildfire perimeters from the CAL FIRE Fire and Resource Assessment Program Historical Fire Perimeters database, downloaded from \url{https://www.fire.ca.gov/what-we-do/fire-resource-assessment-program/fire-perimeters}. MTBS provides severity-based mapping of larger fires, while CAL FIRE includes a broader range of fire sizes. We compared Tessera predictions with two established global burned area products: MCD64A1 Collection 6.1 at 500 m \citep{giglioCollection6MODIS2018a,giglio_justice_boschetti_roy_2021} and GABAM at 30 m \citep{long30ResolutionGlobal2019a,gabamgee}. Both were obtained for California for 2021 from GEE and evaluated using the same reference data and framework described in Section~\ref{sec:wall-to-wall-methods}.

\subsection{European dataset}
\label{sec:european-fires}

To evaluate cross-continental and temporal transfer, we used manually delineated wildfire perimeters from the Copernicus Emergency Management Service Rapid Mapping (EMSR) products for 2024–2025 \citep{emsr}, available at \url{https://mapping.emergency.copernicus.eu/activations/}. We downloaded products for all European fires available for 2025-2025, and when several delineation products were available for the same fire, we retained the latest one. For analyses involving disturbance timing, the earliest post-event satellite acquisition was used as the event date. The resulting dataset contains 88 fires from ten countries: 29 from 2024 and 59 from 2025, with most occurring in fire-prone Mediterranean landscapes in Greece (35), Spain (23), and Portugal (10).

\section{Methods}
\label{sec:methods}

The cohorts, reference labels, and evaluation units used in the primary analyses are summarized in Table~\ref{tab:experimental-overview}.

\begin{table}[t]
\centering
\caption{Primary experimental cohorts and evaluation units. Unit definitions are provided in Section~\ref{sec:hls-burn-scars}.}
\label{tab:experimental-overview}
\small
\renewcommand{\arraystretch}{1.15}
\begin{tabularx}{\textwidth}{@{}p{0.30\textwidth}X@{}}
\toprule
\textbf{Analysis} & \textbf{Cohort and evaluation} \\
\midrule
\textbf{Latent-space analysis}\newline
(Section~\ref{sec:latent-space}) &
1{,}282 eligible fire--chip pairs from 793 chips; balanced burned and unburned samples from refined MTBS masks across $E_{-1}$, $E_0$, and $E_{+1}$; separability scored per pair \\
\addlinespace[4pt]
\textbf{Single-fire mapping}\newline
(Section~\ref{sec:delineation-experiment}) &
Single-fire subset: 559 chips; refined MTBS masks; three-fold spatial CV; full-chip, pixel-pooled scoring \\
\addlinespace[4pt]
\textbf{Annual BA mapping}\newline
(Section~\ref{sec:annual-fire-disturbance}) &
793 chips; union of all same-year refined MTBS burned areas; three-fold spatial CV; full-chip, pixel-pooled scoring \\
\addlinespace[4pt]
\textbf{California wall-to-wall}\newline
(Section~\ref{sec:wall-to-wall-methods}) &
Train: 696 non-California chips with refined MTBS masks. Evaluate: statewide 2021 reference from 37 MTBS fires and 386 CAL FIRE perimeters \\
\addlinespace[4pt]
\textbf{European transfer}\newline
(Section~\ref{sec:cross-continental-transfer}) &
Train: 793 US chips from 2018--2021 with refined annual-fire MTBS labels. Evaluate: 88 European EMSR fire AOIs from 2024--2025, one per fire \\
\addlinespace[4pt]
\textbf{Fire timing}\newline
(Section~\ref{sec:fire-timing}) &
Train: burned pixels from 793 chips using MTBS ignition DOY.\newline
Evaluate US: 1{,}300 eligible fire--chip pairs using three-fold spatial CV; pixel predictions averaged for each pair.\newline
Evaluate Europe: 88 EMSR fires; pixel predictions averaged per fire and compared with the earliest post-event image date. \\
\bottomrule
\end{tabularx}
\end{table}

\subsection{Latent space analysis}
\label{sec:latent-space}

The latent space analysis evaluated whether wildfire disturbance produces a detectable shift in annual embedding space. We assessed whether wildfire increases burned–unburned separability in embedding space by comparing the pre-fire (\(E_{-1}\)), fire-year (\(E_0\)), and post-fire (\(E_{+1}\)) embeddings for both Tessera and AlphaEarth. Burned pixels were sampled from MTBS severity classes 2–4 and unburned pixels from class 0; ambiguous classes 1 and 5 and pixels within 30 m of the refined burned-area boundary were excluded. Unburned pixels were additionally restricted to areas outside all same-year fire perimeters. Embeddings were extracted at their native 10 m resolution.

\textit{Qualitative analysis.} We visualized the structure of the latent space using PCA plots and spatial embedding maps. For the PCA plots, we sampled up to 5,000 pixels per class from each chip and projected \(E_0\) and \(E_{-1}\) embeddings into a joint two-dimensional PCA space, with points coloured by pre-fire land cover or fire-year burn status. For spatial visualization, following \citet{zoranRecurrentVideoMasked2026}, we fitted PCA jointly to \(E_{-1}\) and \(E_0\) and mapped the first three components to RGB channels. In contrast, we fitted unsupervised \(k\)-means clustering (\(k=2\)) separately for each embedding year in the original embedding space and mapped the resulting cluster assignments spatially.

\textit{Quantitative analysis.} We quantified burned--unburned separability at the level of individual fire--chip pairs. For each pair, we randomly sampled up to 1,000 pixels from the target fire and an equal number of unburned pixels outside all same-year fire perimeters. Pairs with fewer than 100 eligible pixels per class were excluded (32 excluded; \(n=1,282\)). For \(E_{-1}\), \(E_0\), and \(E_{+1}\), we calculated (1) cosine distance between burned and unburned class centroids and (2) agreement between burn labels and \(k\)-means clusters (\(k=2\)), allowing cluster labels to be permuted for maximum agreement. Because the classes are balanced, k-means agreement of 0.5 corresponds to random assignment, while 1 indicates perfect separation by burn status.

\subsection{Benchmark experiments}
\label{sec:evaluation-framework}

We used the HLS Burn Scars benchmark to systematically compare annual embeddings with conventional remote sensing imagery for burned area mapping. Each experiment was formulated as a downstream burned area mapping task: the input was either conventional remote sensing imagery or an annual embedding stack, and the output was a binary per-pixel prediction of burned or unburned area. Each input representation was evaluated using the same set of downstream models and fixed training procedures, without hyperparameter tuning, so that performance differences reflect the inputs rather than the models. Choices specific to each experiment, including input years, label, and evaluation, are described below.

\subsubsection{Benchmark setup and evaluation}
\label{sec:evaluation-overview}
\label{sec:reference-labels}
The benchmark setup must account for a mismatch between event-based imagery and annual embedding inputs. Each HLS Burn Scars chip was originally labelled for one target fire. In contrast, Tessera and AlphaEarth embeddings summarize an entire calendar year, so a model operating on these embeddings may respond to any fire that occurred during that year. We therefore used two evaluation protocols. The \emph{single-fire protocol} included only chips in which a single MTBS fire occurred within the chip bounds during the target year. This provides the fairest comparison between annual embeddings and event-specific imagery. The \emph{annual-fire protocol} instead included all benchmark chips, including the ones containing multiple fires that burned in the target year, and evaluated predictions against the union of all same-year fires occurring within each chip. This protocol better matches the temporal support of annual embeddings and tests whether they can identify the chip's entire annual burned area.

\vspace{\baselineskip}
\label{sec:downstream-models}
We used two pixel-level classifiers, L2-regularized logistic regression and Random Forest (RF), and two spatial segmentation models, a standard U-Net and a lightweight U-Net (Table~\ref{tab:downstream_models}). We used logistic regression to test whether burned and unburned pixels were linearly separable in the input feature space, and RF to capture nonlinear pixel-level relationships. We used U-Net-based models to evaluate whether spatial context improves burned area delineation. U-Net is a standard architecture for image segmentation and is commonly used for burned area mapping \citep{ronnebergerUNetConvolutionalNetworks2015a,ribeiroBurnedAreaSemantic2023,seydiBurntNetWildfireBurned2022}. We implemented two variants with different capacities: a lightweight U-Net as a parameter-efficient spatial baseline, and a standard U-Net with a randomly initialized ResNet-18 encoder to test whether performance is limited by downstream model capacity. Full implementation details are provided in Section~\ref{supp:downstream-models} of the Supplementary Material.

\begin{table}[H]
\centering
\caption{Downstream models used in burned area mapping experiments.}
\label{tab:downstream_models}
\small
\begin{tabularx}{\textwidth}{@{}>{\raggedright\arraybackslash}p{2.15cm}>{\raggedright\arraybackslash}p{2.35cm}>{\raggedright\arraybackslash}X>{\raggedright\arraybackslash}X@{}}
\toprule
Model & Type & Key configuration & Training configuration \\
\midrule
Logistic regression &
Linear pixel classifier &
L2 regularization; $C=1$; balanced class weights; standardized inputs &
--- \\
\addlinespace
Random Forest &
Non-linear pixel classifier &
200 trees; $\sqrt{d}$ features/split; unrestricted depth; min.\ leaf 5; min.\ split 10 &
--- \\
\addlinespace
Standard U-Net &
Spatial segmentation &
ResNet-18 encoder, randomly initialized; $\sim$14.5\,M parameters &
BCE + Dice; AdamW ($lr=10^{-4}$); batch 32; max.\ 50 epochs; patience 10 \\
\addlinespace
Lightweight U-Net &
Spatial segmentation &
3-level encoder--decoder; base width 32; $\sim$0.87\,M parameters &
BCE + Dice; AdamW ($lr=10^{-4}$); batch 64; max.\ 50 epochs; patience 10 \\
\bottomrule
\end{tabularx}
\end{table}

\vspace{\baselineskip}
\noindent \label{sec:training}
The inputs used to train downstream models were either conventional HLS imagery or annual embedding stacks from Tessera or AlphaEarth. Pixel-level models were trained using stratified random samples from each training chip, sampling 1,000 burned and 1,000 unburned pixels from the reference mask. Spatial models were trained using eight random $256 \times 256$ crops per chip per epoch, for a maximum of 50 epochs.  

All downstream models were validated using spatially blocked three-fold cross-validation to prevent spatial leakage between nearby chips. Chips were linked into connected components if they shared an MGRS tile, intersected the same MTBS fire event, or had spatial footprints overlapping by more than 1\% of the smaller chip. These components were then partitioned into three approximately equal folds, ensuring that the training and validation sets were disjoint with respect to MGRS tile, fire event, and overlapping footprint. The same folds and training procedure were used for all input representations. For spatial models, a small fixed set of crops from the held-out fold was used to monitor validation loss and select the checkpoint for early stopping (patience 10); no other hyperparameters were tuned using the held-out fold. Final metrics were calculated over the full held-out fold and were reported as mean and standard deviation across the three folds.

Accuracy metrics were standard binary segmentation metrics: precision, recall, F1 score, Intersection over Union (IoU), and overall accuracy. Unless otherwise stated, metrics were reported for the burned class. Predictions were produced on a 10 m grid aligned to the embedding resolution. For benchmark experiments evaluated against MTBS reference labels, the predicted burn probabilities were aggregated from 10 m to 30 m by averaging over each non-overlapping $3\times3$ block of 10 m pixels, and the resulting 30 m probability was thresholded at 0.5 to obtain the 30 m binary prediction.

\subsubsection{Single-fire mapping: comparison with fire-specific HLS imagery}
\label{sec:delineation-experiment}

This experiment tested whether annual embeddings can match event-based burned area delineation without requiring curated fire-specific imagery. We compared four input representations: Tessera \(E_0\), AlphaEarth \(E_0\), post-fire HLS imagery, and paired pre/post HLS imagery. The paired HLS input contained 21 channels: six pre- and six post-fire reflectance bands, pre- and post-fire Normalized Burn Ratio (NBR), differenced NBR (dNBR), and six band-wise reflectance differences. We used the single-fire protocol defined in Section~\ref{sec:reference-labels}, which includes only chips in which a single MTBS fire occurred during the target year. The comparison was restricted to the 559 single-fire chips with more than 80\% valid pixels in both the pre- and post-fire HLS inputs. All four inputs used the same chips, reference masks, and spatially blocked folds. Paired HLS input construction and missing-data handling are detailed in Section~\ref{supp:prefire-hls} of the Supplementary Material.

\subsubsection{Annual burned area mapping}
\label{sec:annual-fire-disturbance}

This experiment tested whether annual embeddings can map the combined burned area of all MTBS fires occurring within each chip during the embedding year. We used all 793 benchmark chips, including those containing multiple fires in the same year, and evaluated Tessera and AlphaEarth fire year embeddings ($E_0$) using logistic regression, Random Forest, and the lightweight U-Net. The reference target was the annual fire mask defined in Section~\ref{sec:reference-labels}, representing the union of all refined MTBS burned area masks for fires occurring within the chip during the corresponding calendar year.

\subsection{Zero-shot regional wall-to-wall mapping}
\label{sec:wall-to-wall-methods}

To test zero-shot regional mapping, we applied a Tessera-trained model wall-to-wall across California in 2021 without using California fire data for training, model tuning or post-processing. We compared its output with GABAM and MCD64A1. The experiment tested whether annual embeddings can map all burned areas across a region without prior knowledge of the region, or where or when the fires occurred. We trained a lightweight U-Net on Tessera \(E_0\) embeddings as input from all 696 benchmark chips outside California using annual-fire labels described in Section~\ref{sec:reference-labels}. Because all available non-California chips were used for training, the model was trained for a fixed 40 epochs rather than using a held-out validation set for early stopping. The trained model was then applied to 2021 Tessera embeddings across California, producing 10 m probabilities thresholded at 0.5 to obtain a binary burned area map. Tessera, GABAM, and MCD64A1 predictions were resampled to the common 30 m MTBS grid using nearest-neighbour resampling.

We performed complementary pixel-level and fire-level analyses. We constructed a statewide reference by combining MTBS and CAL FIRE data. Where MTBS burn-severity data were available, we classified pixels in classes 2–4 as burned; elsewhere, we treated all pixels within CAL FIRE perimeters as burned. For pixel-level evaluation we reported F1, IoU, omission error (\(1-\)recall), and commission error (\(1-\)precision), commonly used for burned area product assessment \citep{padillaValidation2008MODISMCD452014}. Because MTBS severity class 1 (“unburned to low”) is ambiguous, we reported metrics both with class 1 treated as unburned and with it excluded. At the fire level, predicted connected components were compared with 386 individual CAL FIRE fire polygons to quantify detection and spatial agreement across event sizes. A fire was considered detected if any predicted component overlapped it, while spatial agreement for detected fires was assessed separately using fire-level IoU. The same evaluation was applied to Tessera, GABAM, and MCD64A1.

\subsection{Cross-continental transfer}
\label{sec:cross-continental-transfer}

This experiment tested whether fire disturbance representations learned from United States fires generalize across both space and time. We evaluated a model trained on US fires from 2018--2021 on European fires from 2024--2025, providing a cross-continental and cross-year transfer test. We trained the lightweight U-Net model on Tessera $E_0$ embeddings from all 793 chips using refined annual-fire MTBS labels, described in Section~\ref{sec:reference-labels}. The trained model was then applied without retraining to 88 European EMSR fires and evaluated using F1 and IoU. To restrict evaluation to the mapped fire and its immediate surroundings, pixel-level metrics were computed within a rectangle containing a 1~km buffer around each EMSR ground-truth burn perimeter. All pixels inside the resulting axis-aligned bounding box were included in the F1 and IoU calculations.

\subsection{Fire timing prediction}
\label{sec:fire-timing}

We tested whether Tessera and AlphaEarth \(E_0\) embeddings retain information about intra-annual fire timing, using MTBS ignition day of year as the regression target. Because MTBS provides fire-level ignition dates rather than pixel-level burn dates, all burned pixels belonging to the same fire were assigned the same target. This introduces label noise because individual pixels within a fire perimeter may burn days or weeks after ignition. The experiment therefore tested approximate fire-level timing rather than true pixel-level burn timing.

We trained separate Random Forest regression models using Tessera or AlphaEarth \(E_0\) embeddings from reference burned pixels only. The task was therefore conditional on burned area being known: given pixels labelled as burned in the reference data, could the embedding predict when the fire occurred? During inference, the models predicted the day of year for each burned pixel. For the US analysis, pixel-level predictions were averaged within each of the 1{,}300 eligible fire--chip pairs; for the European analysis, they were averaged once per fire. Predictions were evaluated using mean absolute error (MAE), root mean squared error (RMSE), and $R^2$.

We evaluated timing prediction for both Tessera and AlphaEarth in the US benchmark setting, and for Tessera only in the European transfer setting. For the US experiment, we used the same spatially blocked three-fold cross-validation scheme as in Section~\ref{sec:annual-fire-disturbance}. For the European transfer experiment, we trained the regressor on all US fires from the HLS Burn Scars benchmark chips and applied it without retraining to the 88 EMSR fires, using the event dates defined in Section~\ref{sec:european-fires}. Timing performance was additionally stratified by burned-area recall to assess its dependence on successful disturbance detection.

\section{Results}

\subsection{Encoding of wildfire disturbance in annual embeddings}

\begin{figure}[H]
\centering
\includegraphics[width=\textwidth]{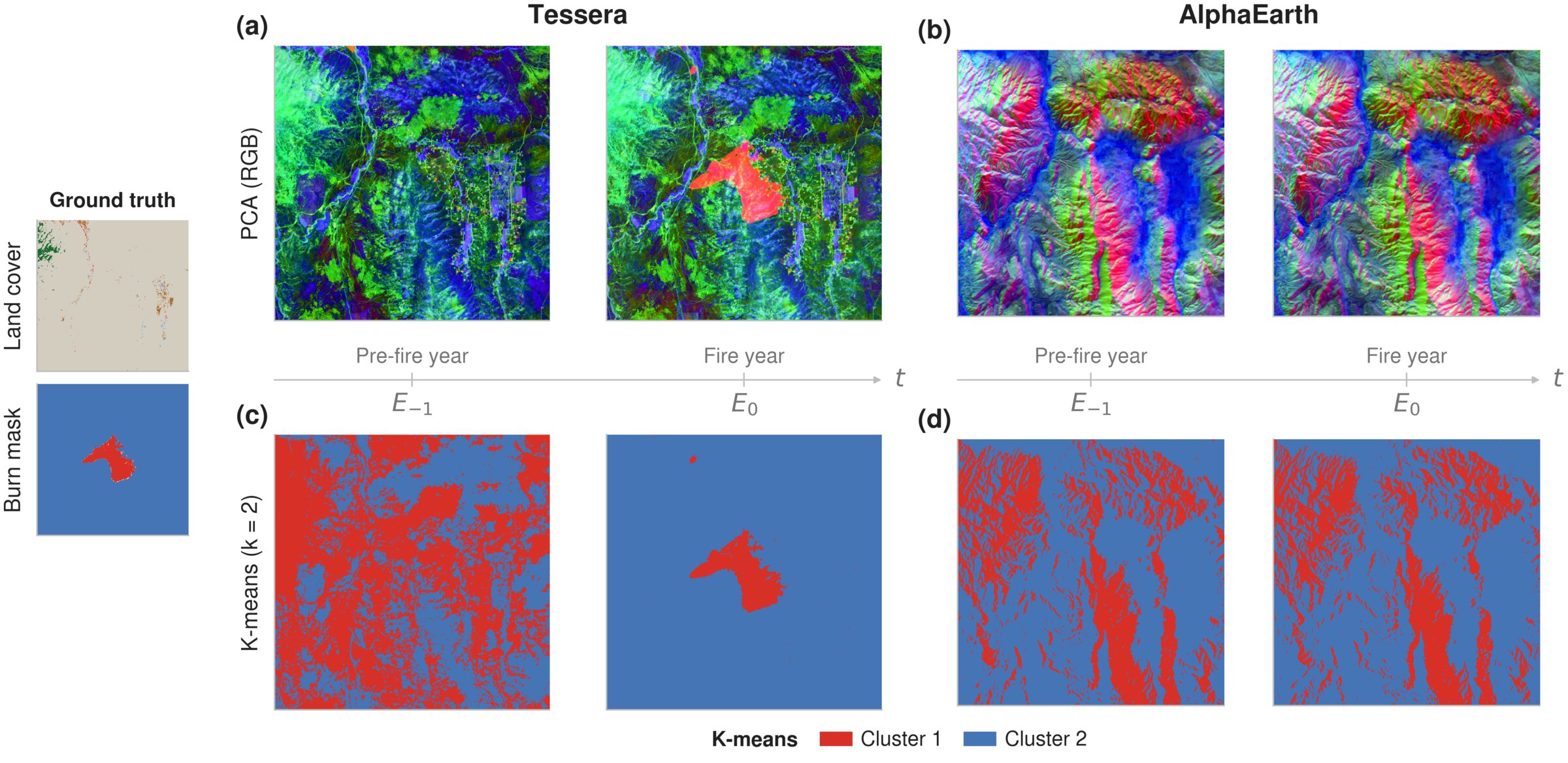}
\caption{Spatial organization of pre-fire and fire year embeddings for an illustrative fire--chip pair. Left panels show pre-fire land cover and the reference burned area. (a--b)~RGB composites of the first three principal components of Tessera and AlphaEarth embeddings for $E_{-1}$ and $E_0$; PCA was fitted jointly across both years for each embedding product. (c--d)~Spatial assignments from unsupervised k-means clustering ($k = 2$), fitted separately to each embedding year.}
\label{fig:latent_qualitative}
\end{figure}

\begin{figure}[H]
\centering
\includegraphics[width=\textwidth]{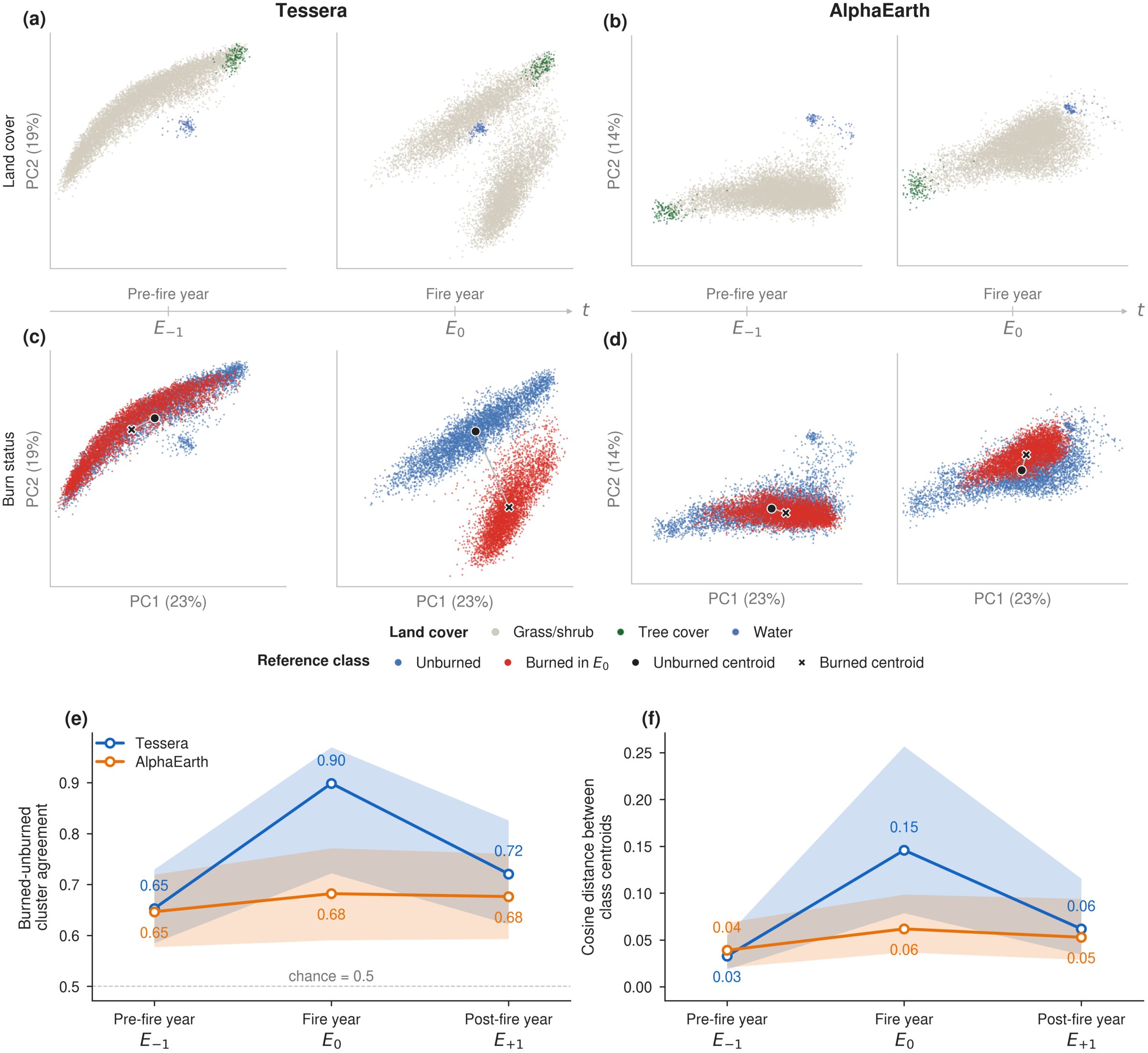}
\caption{Wildfire disturbance reorganizes annual Earth-observation embedding space. (a--d)~PCA projections for an illustrative fire--chip pair, with PCA fitted jointly to ($E_{-1}$) and ($E_0$) embeddings of each product. Points are colored by pre-fire land cover (a--b), or burn status in fire year (c--d). (e--f)~Separability across all eligible fire--chip pairs: (e) agreement between k-means clusters and burn labels, where 0.5 indicates chance agreement; and (f) cosine distance between burned and unburned centroids. Lines show medians and shaded regions the interquartile range.}
\label{fig:latent_quantitative}
\end{figure}

Wildfire disturbance was associated with a marked reorganization of Tessera's latent space during the fire year. Burned and unburned pixels that overlapped substantially in the pre-fire embedding space ($E_{-1}$) became separable in the fire year ($E_0$). This separation weakened again in the post-fire year ($E_{+1}$). The effect is visible in illustrative examples (Figure~\ref{fig:latent_qualitative}; Figure~\ref{fig:latent_quantitative}a--d) and was consistent across the full set of fire--chip pairs (Figure~\ref{fig:latent_quantitative}e--f).

In the absence of disturbance, both Tessera and AlphaEarth showed organization in latent space that is consistent with pre-fire land cover. In $E_{-1}$, PCA projections grouped pixels by land-cover class, with substantial overlap between future burned and unburned samples (Figure~\ref{fig:latent_quantitative}a--b), while the spatial embedding maps primarily reflected landscape structure rather than the future burn perimeter (Figure~\ref{fig:latent_qualitative}). In $E_0$, wildfire became a dominant organizing feature of Tessera embeddings. Burned pixels separated from unburned pixels in the leading PCA dimensions despite variation in pre-fire land cover. The burn scar emerged as a coherent region in RGB composites of the first three principal components, and unsupervised k-means clustering with $k=2$ produced a spatial cluster closely aligned with the burned area, without using burn labels. The corresponding structure was weaker in AlphaEarth, where persistent land-cover variation remained more prominent and k-means did not isolate the burn scar as clearly. These qualitative patterns were common but not universal, with some burn scars showing substantially weaker separation. Additional qualitative examples and nonlinear UMAP projections are provided in Section~\ref{supp:umap} of the Supplementary Material.

Quantitative separability metrics confirmed this pattern across fire–chip pairs (Figure~\ref{fig:latent_quantitative}e--f). For Tessera, k-means agreement with burn labels rose from 0.65 in $E_{-1}$ to 0.90 in $E_0$ before falling back to 0.72 in $E_{+1}$, while cosine distance between burned and unburned centroids increased from 0.03 to 0.15 and returned to 0.06. For AlphaEarth, the corresponding changes were much smaller: k-means agreement moved from 0.65 to 0.68 and 0.68, and centroid distance from 0.04 to 0.06 and 0.05. Wildfire disturbance therefore produced a strong separation between burned and unburned pixels in the fire year Tessera embeddings, but only a marginal shift in AlphaEarth.

\subsection{Annual embeddings delineate burn scars as effectively as event-based approaches}

Models trained on Tessera embeddings matched or exceeded the paired pre- and post-fire HLS baseline for every downstream model, while outperforming post-fire imagery alone and AlphaEarth embeddings (Figure~\ref{fig:compare_with_hls}; Table~\ref{tab:supp_singlefire_metrics} of the Supplementary Material). Tessera with the full U-Net achieved the strongest overall performance (F1 = 0.916, IoU = 0.845), followed by the paired pre/post HLS U-Net (F1 = 0.905, IoU = 0.827). The added temporal context from pre-fire imagery substantially improved the HLS baseline across models (U-Net: F1 = 0.860 to 0.905; RF: 0.679 to 0.798; logistic regression: 0.650 to 0.791). Nevertheless, a single annual Tessera embedding matched the paired fire-specific HLS representation without requiring fire-specific image selection, task-specific feature construction, or explicit spectral differencing. Qualitative maps illustrate these differences across performance regimes (Figure~\ref{fig:singlefire_maps}). The Lions Fire illustrates the importance of acquisition timing for fire-specific inputs: the benchmark HLS image was acquired 30 days after ignition while the fire was still ongoing, capturing only part of the final burn scar.

\begin{figure}[H]
\centering
\includegraphics[width=\textwidth]{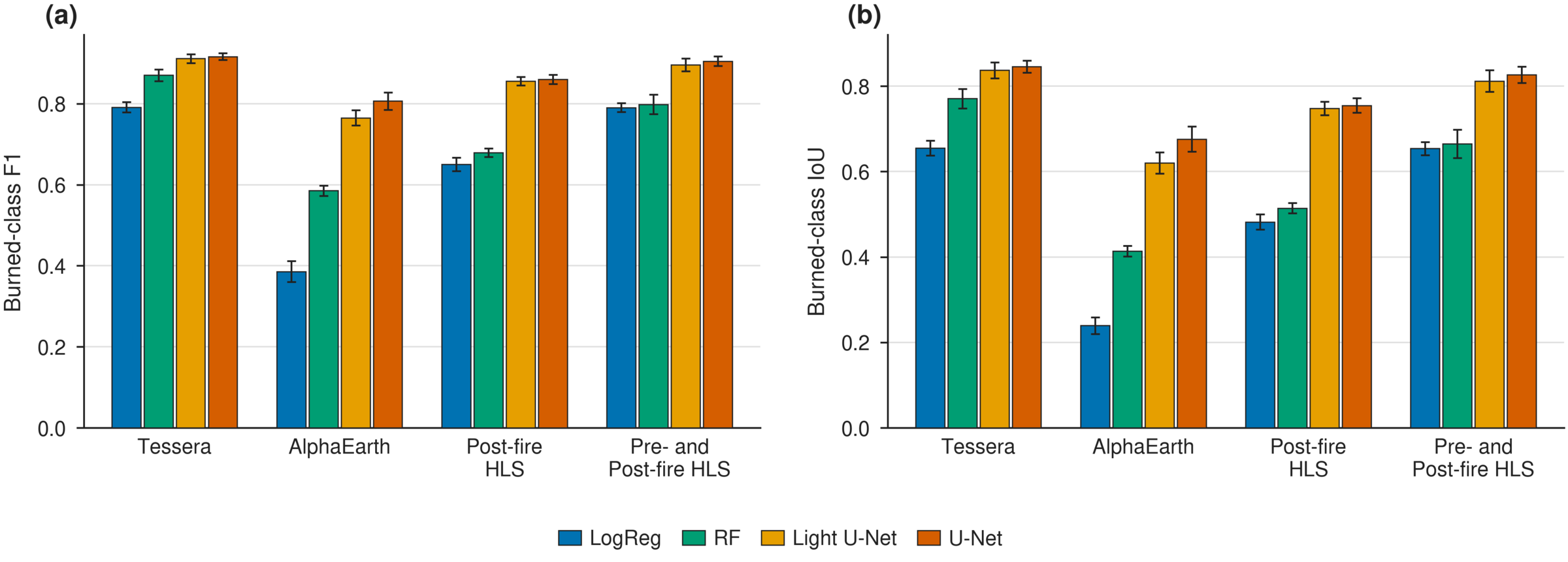}
\caption{Burned area delineation across input representations and downstream models on the 559 single-fire chips. Bars show mean burned-class (a)~F1 and (b)~IoU across three spatially blocked cross-validation folds, and error bars show $\pm$1 standard deviation. Full metrics are provided in Table~\ref{tab:supp_singlefire_metrics} of the Supplementary Material.}
\label{fig:compare_with_hls}
\end{figure}

\begin{figure}[H]
\centering
\includegraphics[width=\textwidth]{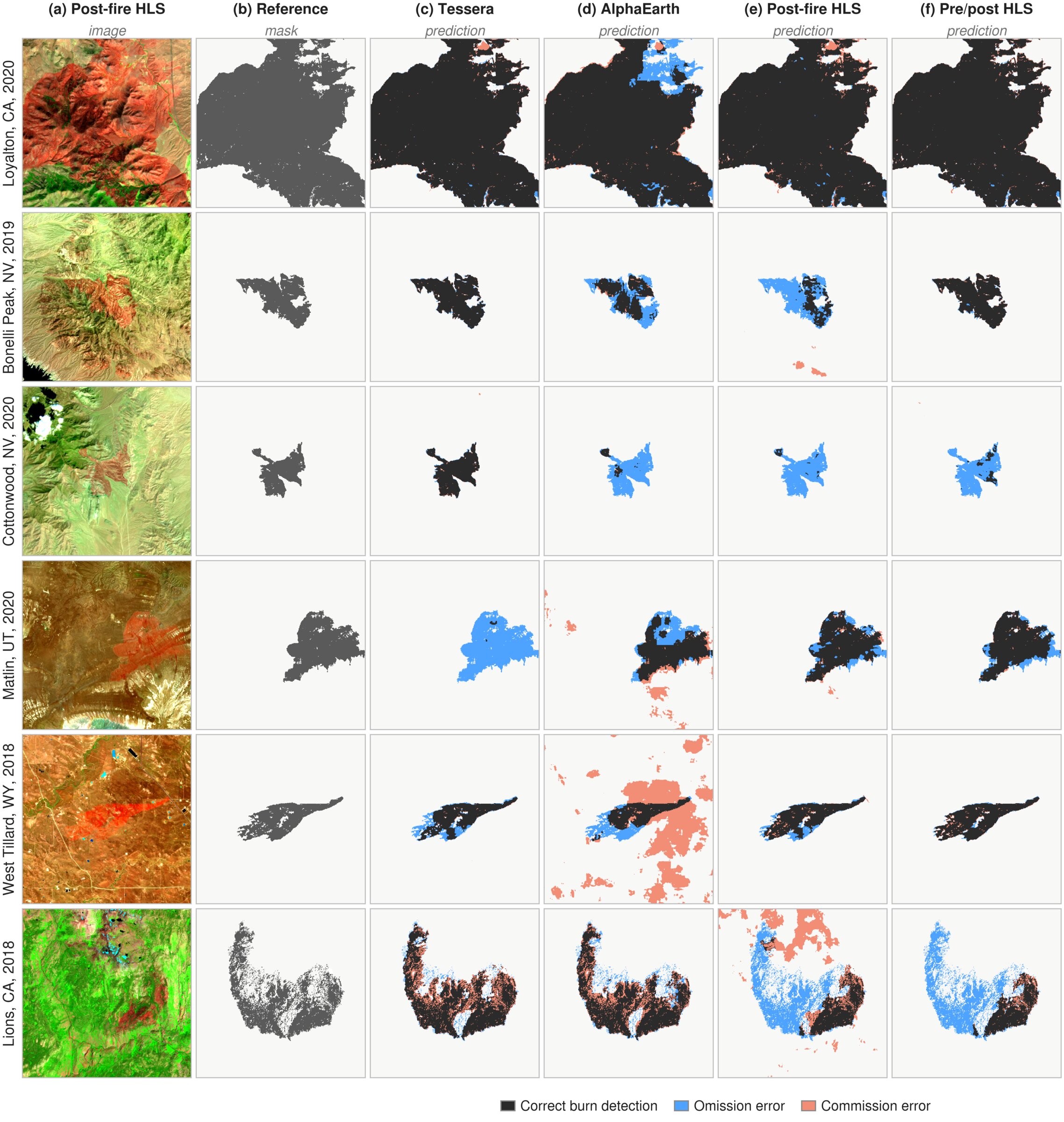}
\caption{Qualitative comparison of single-fire burned area delineation across input representations. Columns show (a)~benchmark post-fire HLS image as a SWIR2--NIR--Red false-colour composite, (b)~the refined MTBS reference, and predictions from (c)~Tessera, (d)~AlphaEarth, (e)~post-fire HLS, and (f)~paired pre/post HLS. Black indicates correct burn detection, blue omission error, and orange commission error. Rows were selected to illustrate characteristic performance regimes across approaches.}
\label{fig:singlefire_maps}
\end{figure}

The effect of downstream model choice depended strongly on the input representation. Tessera already performed strongly with pixel-level models: logistic regression reached F1 = 0.791, matching the paired HLS baseline, while the Tessera RF reached F1 = 0.870 and slightly outperformed the full U-Net trained on post-fire HLS alone (F1 = 0.860), despite using no spatial context. Adding spatial modelling provided much larger gains for representations with weaker pixel-level performance. Moving from RF to the lightweight U-Net increased F1 by 0.18 for both AlphaEarth and post-fire HLS, compared with 0.10 for paired HLS and only 0.04 for Tessera. Notably, AlphaEarth showed the largest gain despite its embeddings already containing patch-level spatial context. Increasing U-Net capacity provided little additional benefit: moving from the lightweight U-Net (0.87 M parameters) to the full U-Net (14.5 M) changed F1 by less than 0.01 for Tessera and both HLS inputs, and by 0.04 for AlphaEarth. In this experiment, the choice of input representation mattered far more than the capacity of the downstream model.

\subsection{Annual embeddings accurately map the combined burned area of multiple fires}
\label{sec:annual-fire-mapping-results}

Tessera maintained strong performance when mapping combined burned area from multiple MTBS fires within each chip and embedding year, rather than only the benchmark target fire. Across 793 chips containing 1,162 distinct MTBS fires, the Tessera lightweight U-Net achieved F1 = 0.896 $\pm$ 0.016 and IoU = 0.813 $\pm$ 0.027 across the three cross-validation folds (Table~\ref{tab:all_fires_summary}).

\begin{table}[H]
\centering
\caption{Burned area mapping performance across 793 benchmark chips containing 1{,}162 distinct MTBS fires. Results are mean $\pm$ standard deviation across three spatially blocked cross-validation folds.}
\label{tab:all_fires_summary}
\begin{tabular}{llcc}
\toprule
\textbf{Representation} & \textbf{Model} & \textbf{IoU $\uparrow$} & \textbf{F1 $\uparrow$} \\
\midrule
Tessera     & LogReg      & $0.642 \pm 0.041$ & $0.781 \pm 0.031$ \\
Tessera     & RF          & $0.739 \pm 0.038$ & $0.849 \pm 0.025$ \\
\textbf{Tessera} & \textbf{Light U-Net} & $\mathbf{0.813 \pm 0.027}$ & $\mathbf{0.896 \pm 0.016}$ \\
\midrule
AlphaEarth  & LogReg      & $0.241 \pm 0.033$ & $0.388 \pm 0.043$ \\
AlphaEarth  & RF          & $0.402 \pm 0.053$ & $0.571 \pm 0.054$ \\
AlphaEarth  & Light U-Net & $0.602 \pm 0.044$ & $0.751 \pm 0.035$ \\
\bottomrule
\end{tabular}
\end{table}

\begin{figure}[H]
\centering
\includegraphics[width=\textwidth]{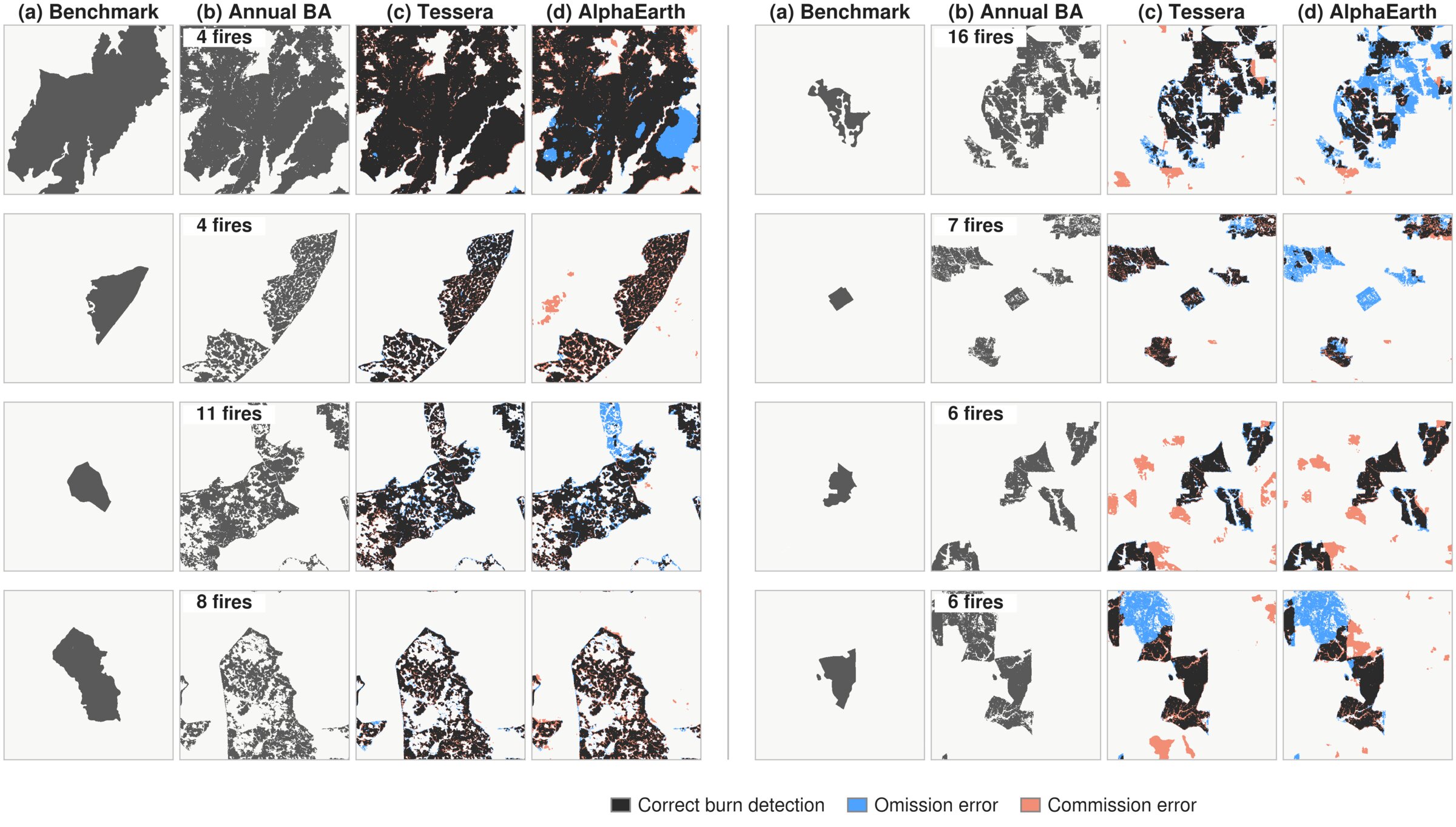}
\caption{Examples of multi-fire benchmark chips. Columns show (a)~the original target-fire label, (b)~the annual reference containing all same-year fires, and predictions from (c)~Tessera and (d)~AlphaEarth. Numbers in (b) indicate the number of same-year MTBS fires intersecting each chip. Dark grey indicates correct burn detection, blue omission error, and red commission error.}
\label{fig:annual_fires}
\end{figure}

Annual embeddings also recovered fires beyond the original benchmark target fire. In multi-fire chips, Tessera frequently delineated additional same-year fires present in the annual reference but absent from the original benchmark label (Figure~\ref{fig:annual_fires}). Such detections would be counted as false positives under the original single-fire benchmark but are correct under the annual-fire protocol. Despite the broader prediction target, Tessera lightweight U-Net performance declined only slightly relative to the single-fire experiment, from F1 = 0.911 to 0.896.

Tessera outperformed AlphaEarth across all downstream models, with the largest gap occurring for the simplest models. The narrowing gap with increasing model capacity is consistent with the latent-space results: Tessera embeddings provide a more directly separable burn signal, while a spatial decoder partially compensates for weaker separability in AlphaEarth. For AlphaEarth, the gain from spatial modeling was driven mainly by precision, which increased from 0.260 with logistic regression to 0.733 with the lightweight U-Net, while recall remained almost unchanged (0.770 to 0.772). Spatial context therefore primarily suppressed false positives rather than recovering many additional burned pixels.

\begin{figure}[H]
\centering
\includegraphics[width=\textwidth]{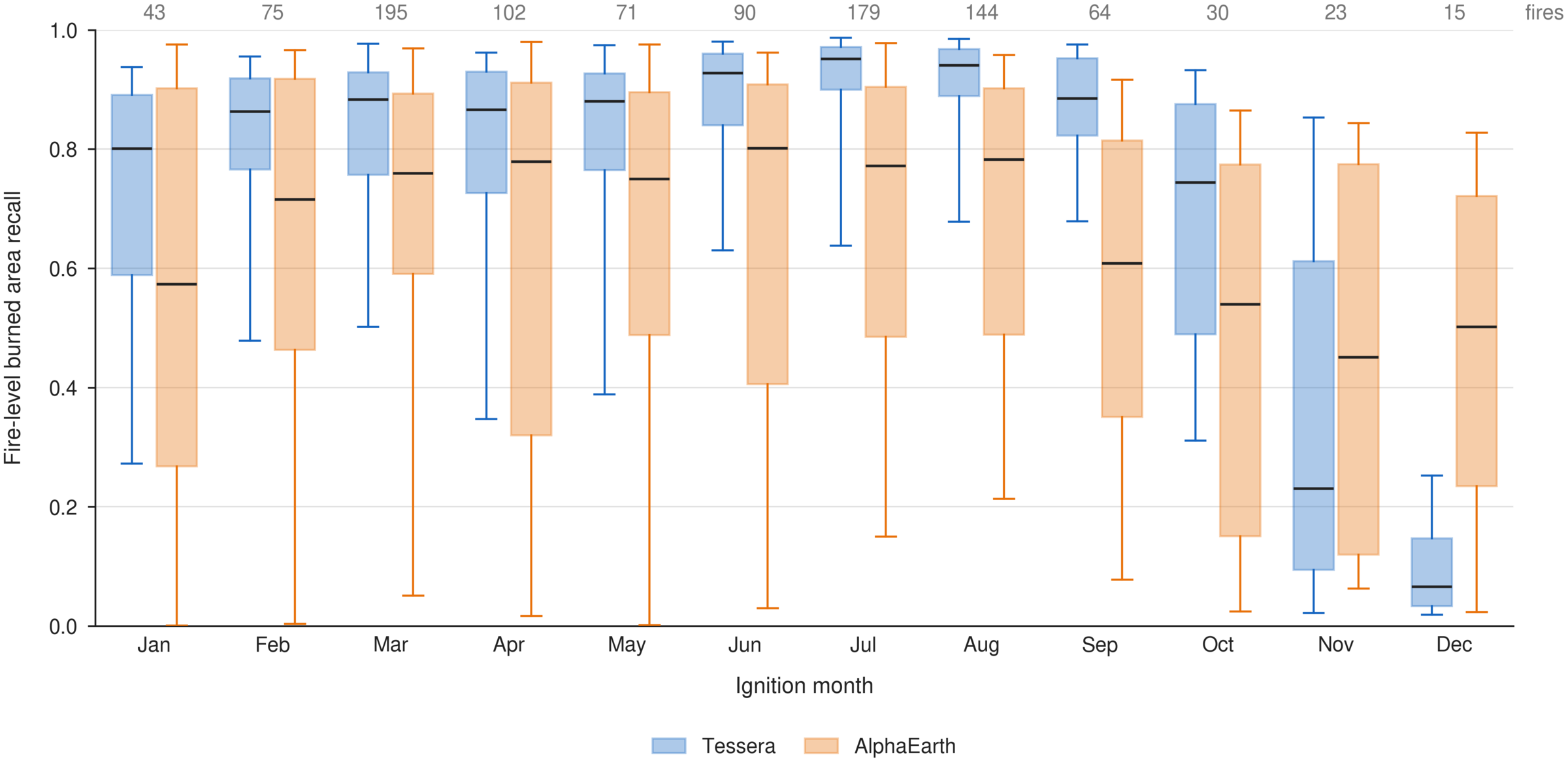}
\caption{Fire-level burned area recall by ignition month for Random Forest models using Tessera and AlphaEarth fire-year embeddings. Boxplots show recall across MTBS fire events; numbers indicate fires per month.}
\label{fig:limitations}
\end{figure}

Fire-level recall varied with ignition timing (Figure~\ref{fig:limitations}). Tessera recall remained high through most of the year but declined sharply for fires ignited near the end of the calendar year, particularly in November and December. AlphaEarth showed lower and more variable recall throughout the year and a weaker raw late-season decline. Covariate-adjusted analyses nevertheless confirmed reduced late-season recall for both representations after accounting for fire and landscape characteristics (Section~\ref{supp:late-season} of the Supplementary Material).

 \subsection{Tessera supports zero-shot regional mapping in California}

\begin{figure}[H]
 \centering
 \includegraphics[width=\textwidth]{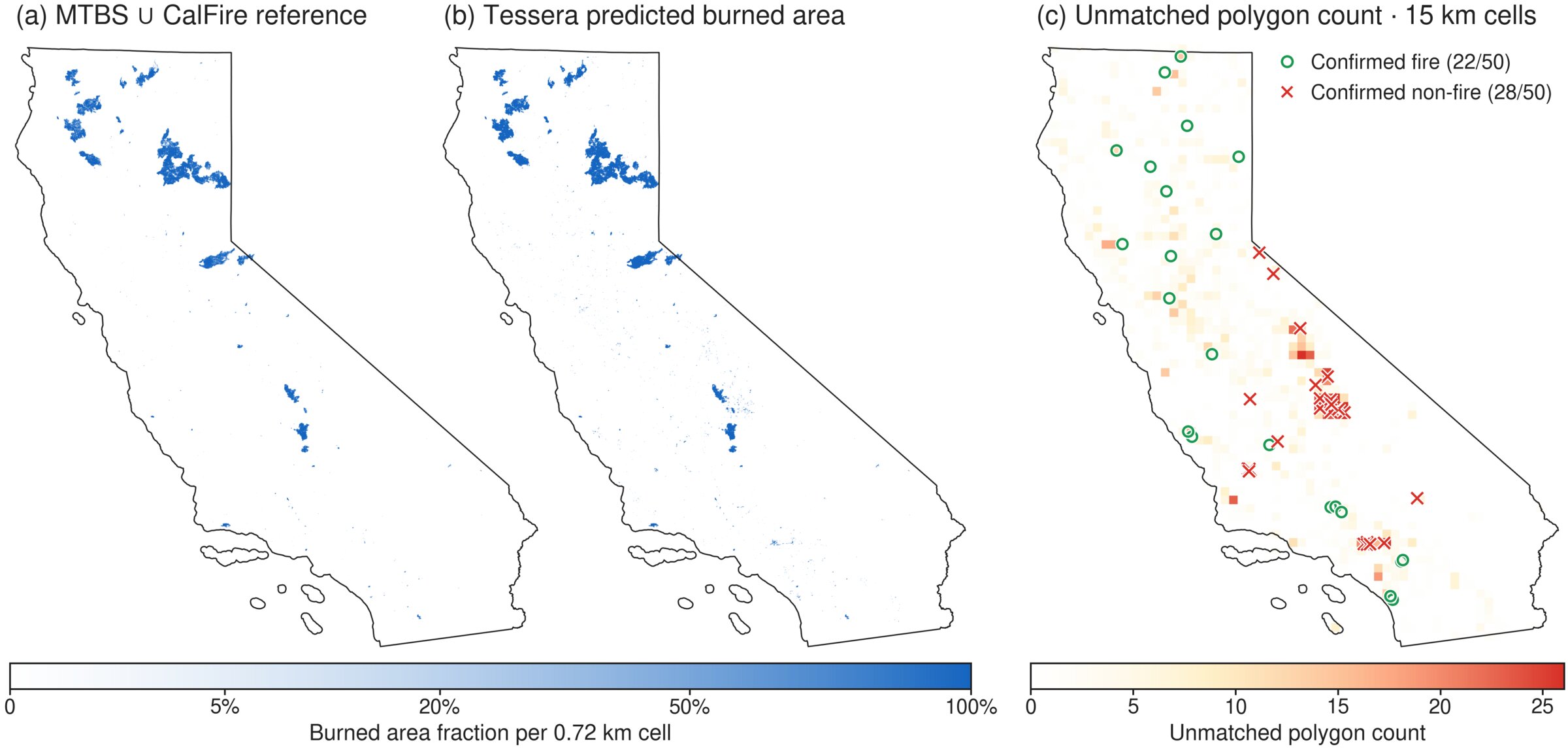}
 \caption{Zero-shot wall-to-wall burned area mapping across California in 2021. (a)~Combined MTBS and CAL FIRE burned area reference and (b)~Tessera prediction, shown as the fraction of burned area within 0.72~km grid cells. A nonlinear color normalization is used in (a--b) to improve visibility of low burned area fractions at statewide scale. (c)~Spatial distribution of Tessera prediction polygons unmatched by either MTBS or CAL FIRE, shown as counts per 15~km grid cell after excluding polygons within 300~m of a reference perimeter. Symbols show the 50 largest manually reviewed unmatched polygons: 22 were confirmed as genuine fires absent from both reference inventories and 28 as non-fire detections.}
 \label{fig:unmatched_overview}
 \end{figure}

 A Tessera model trained entirely outside California successfully mapped statewide burned area for 2021 (Figure~\ref{fig:unmatched_overview}(a,b)). This deployment setting is deliberately demanding: the model was trained only on benchmark chips, with no statewide negative examples and no region-specific post-processing. The model recovered nearly all reference burned area while also detecting more small and medium-sized fires than the GABAM and MCD64A1 products. Its main apparent weakness is the number of potential false positives, but manual inspection showed that many of the largest unmatched predictions were real fires absent from both reference datasets.
 
 \begin{table}[H]
 \centering
 \caption{Statewide pixel-level performance of burned area products in California in 2021 against the combined MTBS--CAL FIRE reference. Reference data comprise 37 MTBS fires and 386 CAL FIRE fire perimeters. Metrics are reported with MTBS severity class 1 (``unburned to low'') either treated as unburned or excluded from evaluation. Omission error (OE) is unchanged between treatments and is therefore reported once.}
 \label{tab:california_pixel_metrics_combined}
 \begin{tabular}{l c ccc ccc}
 \toprule
 Product &
  &
 \multicolumn{3}{c}{Severity 1 as unburned} &
 \multicolumn{3}{c}{Severity 1 excluded} \\
 \cmidrule(lr){3-5}
 \cmidrule(lr){6-8}
 &
 OE $\downarrow$ &
 CE $\downarrow$ &
 F1 $\uparrow$ &
 IoU $\uparrow$ &
 CE $\downarrow$ &
 F1 $\uparrow$ &
 IoU $\uparrow$ \\
 \midrule
 Tessera &
 \textbf{3.0\%} &
 14.6\% &
 \textbf{0.908} &
 \textbf{0.831} &
 5.9\% &
 \textbf{0.955} &
 \textbf{0.914} \\
 MCD64A1 &
 11.8\% &
 15.8\% &
 0.862 &
 0.757 &
 6.4\% &
 0.908 &
 0.832 \\
 GABAM &
 45.8\% &
 \textbf{7.0\%} &
 0.684 &
 0.520 &
 \textbf{4.3\%} &
 0.692 &
 0.529 \\
 \bottomrule
 \end{tabular}
 \end{table}
 
 According to the pixel-level metrics in Table~\ref{tab:california_pixel_metrics_combined}, the Tessera model achieved the lowest omission error at 3.0\%, compared with 11.8\% for MCD64A1 and 45.8\% for GABAM. Commission error was comparable between Tessera and MCD64A1, at 14.6\% and 15.8\%, respectively. GABAM's commission was substantially lower at 7.0\%, but its conservative predictions came at the cost of missing almost half of the reference burned area. Much of the apparent commission error for Tessera and MCD64A1 occurred within pixels labeled as MTBS severity class 1 in the 37 large fires mapped by MTBS. Severity 1 is defined as ``unburned to low severity'', so the burn status of these pixels is ambiguous in the reference dataset itself. By default, we treated these pixels as unburned, consistent with the labels used for model training. With severity 1 pixels excluded from evaluation, commission errors dropped for all three products, as shown in Table~\ref{tab:california_pixel_metrics_combined}, with Tessera's and MCD64A1's errors falling by more than half. Excluding severity 1 substantially reduced apparent commission, but did not change the ranking or overall interpretation of the products.
 
 \begin{figure}[H]
 \centering
 \includegraphics[width=\textwidth]{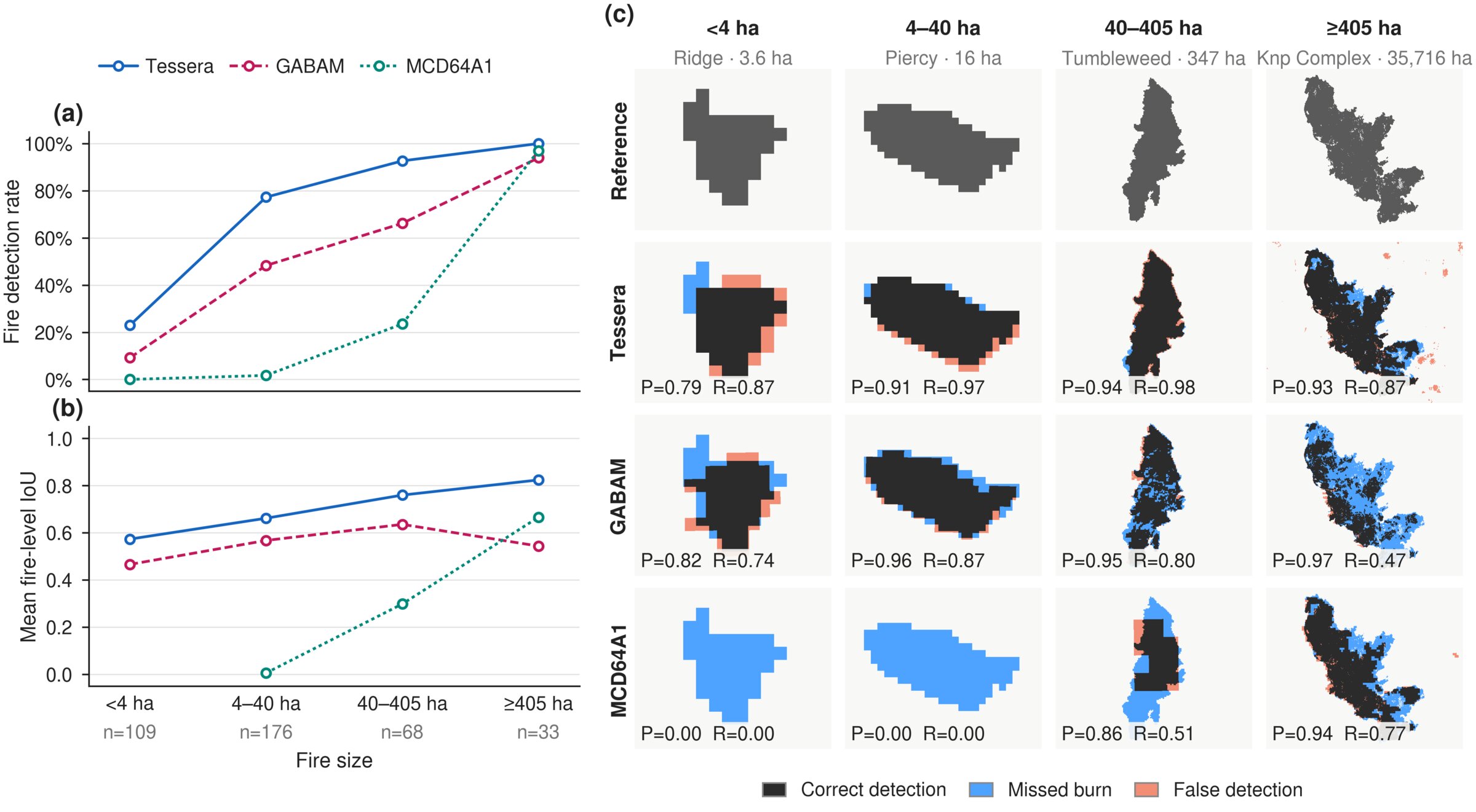}
 \caption{Fire detection and delineation across fire sizes in California in 2021. (a)~Detection rate for 386 CAL FIRE fires by size class; (b)~mean fire-level IoU among detected fires; and (c)~representative examples comparing the reference with Tessera, GABAM, and MCD64A1. Black indicates correct burn detection, blue omission, and orange commission; $P$ and $R$ denote precision and recall.}
 \label{fig:fire_size_response}
 \end{figure}
 
 Pixel-level metrics were dominated by large fires that represent most statewide burned area. Fire-level evaluation revealed a larger advantage for Tessera at smaller fire sizes (Figure~\ref{fig:fire_size_response}). For fires above 405~ha, all three products had similarly high detection rates, but Tessera detected 92.7\% of fires between 40 and 405 ha, compared with 66.2\% for GABAM and 23.5\% for MCD64A1. For fires between 4 and 40 ha, detection remained 77.3\% for Tessera, compared with 48.3\% and 1.7\%, respectively. MCD64A1 was effectively insensitive to fires below 40~ha, and all three products struggled below 4~ha, although Tessera retained the highest detection rate (22.9\%). Among detected fires, Tessera predictions also showed greater spatial agreement with the reference perimeters, achieving the highest mean fire-level IoU in every size class (Figure~\ref{fig:fire_size_response}(b)).
 
 Tessera produced 2,713 predicted components that did not intersect any MTBS or CAL FIRE perimeter, compared with 820 for GABAM and 170 for MCD64A1. Despite their large number, these components accounted for only a small share of each product's total predicted burned area: 4.7\% for Tessera, 4.0\% for GABAM, and 2.4\% for MCD64A1. Most unmatched Tessera components were small, although components between 40 and 405~ha accounted for the largest share of unmatched area. Reference disagreement did not always indicate model error: manual inspection of the 50 largest unmatched Tessera components found that 22 were genuine fires visible in satellite imagery but absent from both reference datasets. Confirmed fires were dispersed across California, whereas confirmed non-fire detections clustered in the snow-covered, high-elevation southern Sierra Nevada and the hills surrounding Los Angeles, with additional errors in agricultural regions (Figure 8c). The broader unmatched-polygon distribution showed similar spatial concentrations, indicating that commission errors are systematic rather than random.
 
\subsection{Cross-continental transfer}

\begin{figure}[H]
\centering
\includegraphics[width=\textwidth]{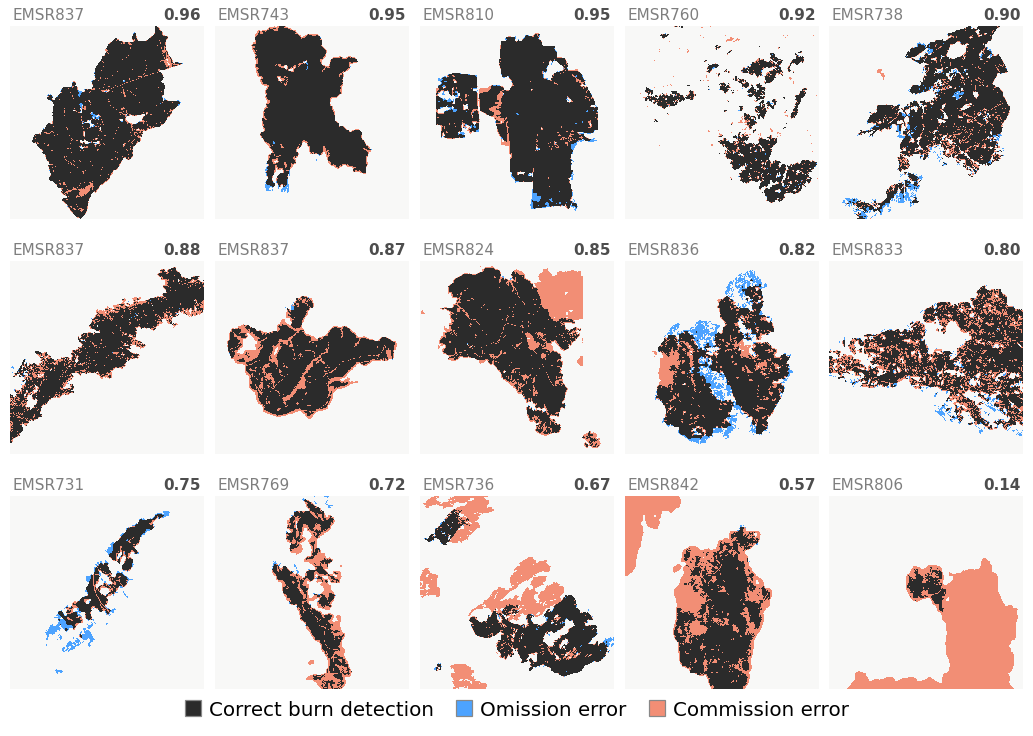}
\caption{Per-fire Tessera lightweight U-Net predictions for 15 European EMSR fires from 2024--2025. Examples span the observed performance range and are ordered by decreasing IoU. Black indicates correct burn detection, blue omission, and orange commission error; bold values show per-fire IoU.}
\label{fig:spatial_transfer}
\end{figure}

Burned area segmentation remained strong under transfer across continents, reference data sources, and evaluation years. A light U-Net trained on US HLS Burn Scars benchmark chips from 2018--2021 was applied without retraining to 88 European EMSR fires from 2024--2025, achieving micro F1 = 0.882 and IoU = 0.789 (Table~\ref{tab:spatial_transfer}). Performance was stable across evaluation years, with F1 = 0.881 in 2024 and 0.883 in 2025. Fire-level macro scores were slightly lower (macro F1 = 0.829 $\pm$ 0.142), with most fires mapped accurately while a small number of events performed substantially worse (Figure~\ref{fig:spatial_transfer}).

\begin{table}[H]
\centering
\caption{Cross-continental burned area segmentation on 88 European EMSR fires from 2024--2025 using a Tessera model trained on US fires from 2018--2021. Micro metrics aggregate pixels across fires; macro metrics are mean $\pm$ standard deviation across fires.}
\label{tab:spatial_transfer}
\begin{tabular}{lrrrrr}
\toprule
Evaluation set & $n$ fires & Micro F1 $\uparrow$ & Micro IoU $\uparrow$ & Macro F1 $\uparrow$ & Macro IoU $\uparrow$ \\
\midrule
2024 & 29 & 0.881 & 0.787 & $0.849 \pm 0.105$ & $0.750 \pm 0.137$ \\
2025 & 59 & 0.883 & 0.790 & $0.820 \pm 0.157$ & $0.718 \pm 0.180$ \\
Combined & 88 & 0.882 & 0.789 & $0.829 \pm 0.142$ & $0.728 \pm 0.168$ \\
\bottomrule
\end{tabular}
\end{table}

\begin{figure}[H]
\centering
\includegraphics[width=\textwidth]{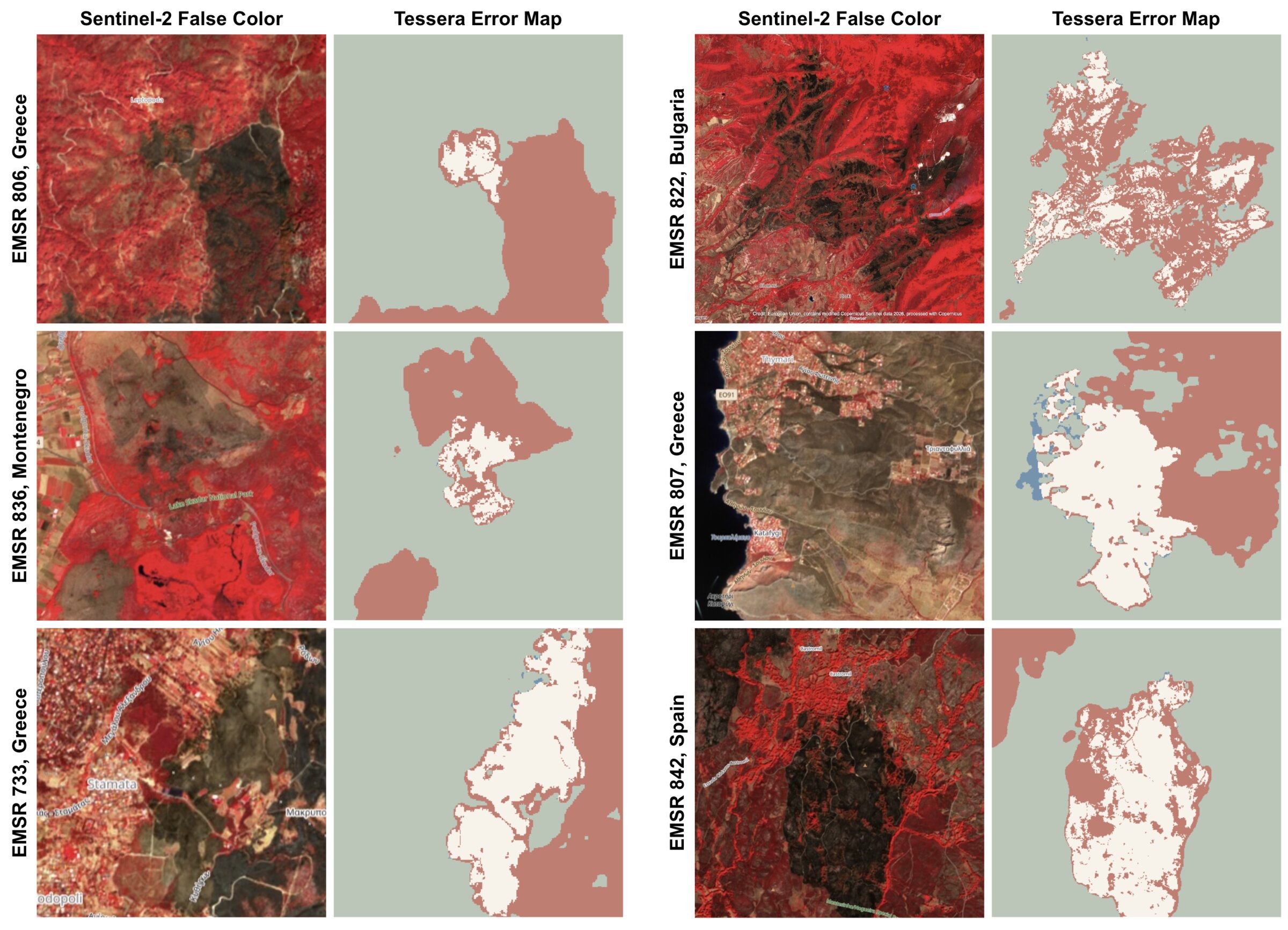}
\caption{Qualitative error analysis for selected low-performing European EMSR fires. For each event, a Sentinel-2 false color composite (left) is paired with a Tessera Light U-Net error map (right): true positives (white), true negatives (light green), false positives (red), and false negatives (blue). Many false positives (red) coincide with visible burn scars outside the EMSR evaluation perimeter.}
\label{fig:false_colors}
\end{figure}

Inspection of poorly mapped fires showed that many apparent false positives corresponded to actual burned areas from nearby fires that occurred in the same year. False color Sentinel-2 imagery for the six fires that scored F1 below 0.6 (Figure~\ref{fig:false_colors}) revealed the additional burned area around the reference EMSR perimeters that matched Tessera's predicted burned area. These detections were counted as false positives, reflecting the mismatch between annual embeddings and single-fire reference labels discussed in Section~\ref{sec:annual-fire-mapping-results}. The reported transfer metrics are therefore conservative, and the worst-scoring fires would rank considerably higher against an all-fires annual reference.

\subsection{Tessera embeddings capture within-year timing of fires}

\begin{figure}[H]
\centering
\includegraphics[width=\textwidth]{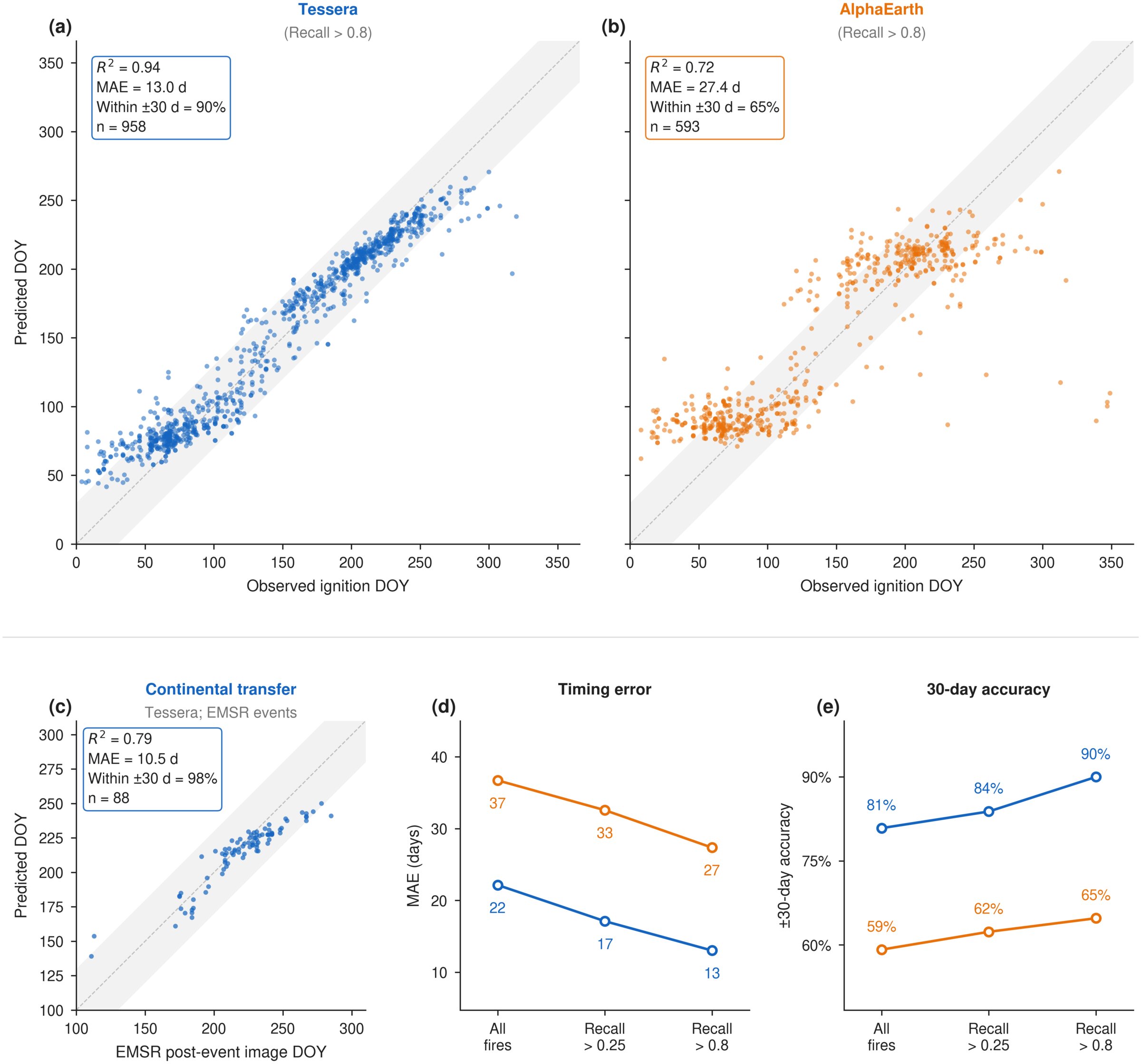}
\caption{Fire timing prediction from annual embeddings. (a--b)~Predicted vs observed ignition day of year for Tessera and AlphaEarth on fires with burned area recall $>0.8$. (c)~Continental transfer of Tessera timing prediction to European EMSR events. (d--e)~Mean absolute error and $\pm$30 day accuracy as a function of burn detection recall threshold.}
\label{fig:timing}
\end{figure}

Annual Tessera embeddings contain information about when a fire occurred within the year, not only whether it occurred. Timing accuracy depended strongly on how well the burned area was detected (Figure~\ref{fig:timing}(d)). Across all fires, Tessera predicted ignition day of year with MAE = 22 days and dated 81\% of fires within $\pm$30 days. Performance improved to MAE = 17 days for fires with burned area recall $>0.25$ and to MAE = 13.0 days ($R^2$ = 0.94; 90\% within $\pm$30 days; $n$ = 958) for fires with recall $>0.8$. AlphaEarth also retained some within-year timing information, but with lower accuracy: across all fires, MAE was 37 days and 59\% were dated within $\pm$30 days; among well-detected fires (recall $>0.8$), MAE improved to 27.4 days ($R^2$ = 0.72; 65\% within $\pm$30 days; $n$ = 593). Fires that were poorly detected spatially were therefore also more difficult to date for both representations, consistent with both tasks depending on how strongly the disturbance is represented in the annual embedding.

Timing prediction also transferred to European fires (Figure~\ref{fig:timing}(c)): a model trained only on US fires dated 98\% of 2024--2025 EMSR fires within $\pm$30 days of the reported event date, with MAE = 10.5 days ($R^2$ = 0.79; $n$ = 88). Because the European fires occupy a narrower seasonal window than the US benchmark fires, $R^2$ values are not directly comparable across regions, but the low absolute errors indicate that the encoded timing information is not specific to US fire regimes. Thus, although Tessera embeddings are annual summaries, they retain sufficient temporal structure to estimate approximate ignition timing for detected fires, extending their utility beyond burned area delineation.

\section{Discussion}

\subsection{Wildfire disturbance is encoded within a single annual embedding}

A central result of this study is that wildfire disturbance is encoded directly within a single annual embedding. Fire year Tessera embeddings for burned pixels separated clearly from unburned ones in latent space, and downstream models trained only on the fire year embedding recovered burned area with high accuracy. Even pixel-level logistic regression performed strongly, with no spatial context and no temporal differencing. The disturbance signal is already present in the representation itself. Such information is plausible given what a spectral-temporal embedding is: a summary of a pixel's full yearly trajectory. Unless a fire occurs at the calendar boundary, the annual observation sequence contains both pre- and post-fire imagery, and the trajectory of a burned pixel looks fundamentally different from that of an undisturbed one. A model trained to summarize trajectories can therefore learn to express the change itself. In this sense, annual embeddings act as a learned representation of the incomplete 2D spectral-temporal surface (DOY $\times$ wavelength). Conventional composites also summarize a year of observations, but through fixed rules such as medians, best-available pixels, or spectral indices, which flatten temporal structure and can lose short-lived disturbance signals. Learned spectral-temporal embeddings instead preserve whatever aspects of the trajectory are useful for downstream tasks, including whether that trajectory was stable, followed a phenological pattern, or was abruptly disturbed.

Change detection is often formulated as an explicit comparison between pre- and post-event observations. The European Forest Disturbance Atlas compares seasonal Landsat composites of consecutive years \citep{viana-sotoEuropeanForestDisturbance2025a}, CCDC fits harmonic models to the pre-disturbance period of spectral bands and searches for breaks from the fitted trajectory \citep{zhuContinuousChangeDetection2014a}, and global burned area products traverse dense time series looking for temporal changes in burn-sensitive spectral information \citep{giglioCollection6MODIS2018a,long30ResolutionGlobal2019a}. Post-fire segmentation approaches instead classify burn scars from a single well-timed image \citep{huUniTemporalMultispectralImagery2021,seydiBurntNetWildfireBurned2022,fuBurnedAreaSegmentation2024}, relying on the persistence of the spectral fire signal rather than on change itself. Studies that add pre-fire imagery \citep{suiBiAUNetWildfireBurnt2024}, together with our own benchmarks, show that paired pre- and post-fire inputs outperform post-fire imagery alone, highlighting the value of explicit temporal contrast.

Existing approaches thus either measure change explicitly between observations or forgo temporal information and classify its lasting spectral imprint. Temporal embeddings represent a conceptual shift: the comparison is internalized in the representation and moved upstream into foundation model pre-training. The downstream model neither compares observations nor depends on a well-timed post-fire snapshot; it reads out a single learned summary in which the disturbance is already encoded.

Most current geospatial foundation models are not temporal descriptors \citep{fengTESSERATemporalEmbeddings2026}: they encode observations from a single time step, so disturbance mapping with them still requires either explicit pre/post comparison or classification of the post-event state alone. The HLS Burn Scars results illustrate the limitation of the latter. In PANGAEA, a supervised U-Net trained directly on post-fire imagery outperformed all tested geospatial foundation models encoding the same post-fire input \citep{marsocciPANGAEAGlobalInclusive2024}. In our contextual comparison, even a Random Forest on a single annual Tessera embedding exceeded the published post-fire U-Net result, although the evaluation protocols are not strictly head-to-head (Section~\ref{supp:pangaea} of the Supplementary Material). The distinction is that Tessera summarizes the temporal trajectory containing both pre- and post-fire observations, rather than learning a representation of the post-fire state alone. Notably, even AlphaEarth, which summarizes observations over time, evaluates change by explicitly comparing sequential annual embeddings, concatenating the before- and after-period representations for supervised change classification \citep{brownAlphaEarthFoundationsEmbedding2025}. \citet{seydiDeepLearningBasedBurned2025a} adopted this approach for burned area mapping, using paired pre- and post-fire AlphaEarth embeddings in a bi-temporal Siamese U-Net with explicit feature differencing. In a contextual comparison on the same 17-fire EMSR cohort, we found that the lightweight U-Net trained on a single annual Tessera embedding achieved higher F1 and IoU than this bi-temporal AlphaEarth approach (Section~\ref{supp:seydi} of the Supplementary Material). For Tessera, by contrast, adding explicit temporal context from adjacent embedding years yielded little additional improvement (Section~\ref{supp:multiyear} of the Supplementary Material), suggesting that a single annual embedding whose summary window contains the disturbance is often sufficient.

\subsection{Disturbance sensitivity differs across foundation models}

Tessera and AlphaEarth both produce annual temporal embeddings and both incorporate Sentinel-1 and Sentinel-2 observations, yet they behave very differently for wildfire disturbance, in both latent structure and downstream accuracy. In Tessera, the fire year embedding space is strongly reorganized by burn state: burned and unburned pixels separate clearly, simple classifiers perform well, and spatial models need only modest capacity to reach the best results. AlphaEarth embeddings show much weaker burned and unburned separability, and downstream models require substantially more spatial modeling to reach their best performance. This divergence is not specific to our setting. A concurrent, independent evaluation of both embedding products for burned area classification in Portugal, using different reference perimeters and pixel-level classifiers, likewise found Tessera consistently stronger than AlphaEarth \citep{silvaEvaluatingAlphaEarthTESSERA2026}, and on the 17 European fires common to our study and that of \citet{seydiDeepLearningBasedBurned2025a}, a single-year Tessera model exceeded their bi-temporal AlphaEarth network (Section~\ref{supp:seydi} of the Supplementary Material). In contrast, the two models perform comparably on many other tasks \citep{ballGeospatialFoundationModels2026}, with AlphaEarth performing better on some \citep{liuCITYREPUnifiedBenchmark2026,plasBetterTogetherEvaluating2026}. Most of those benchmarks target environmental states or properties, such as land cover, vegetation structure, biomass, or crop type. Burned area mapping instead asks whether the representation preserves a discrete disturbance event within an annual summary, and for this task the two models diverge sharply.

Both representations capture high-severity burns well, but the gap widens for lower-severity burns. Using out-of-fold Lightweight U-Net predictions across all 793 benchmark chips, false-negative rates for high-severity pixels were 0.1\% for Tessera and 0.8\% for AlphaEarth, but rose to 10.8\% and 29.0\%, respectively, for low-severity pixels (Table~\ref{tab:supp_severity_fnr} of the Supplementary Material). AlphaEarth therefore appears able to identify high-severity burned areas reliably, but struggles increasingly as burn severity decreases. Because low-severity pixels make up the majority of burned area in our benchmark, this weakness likely contributes substantially to its lower aggregate performance.

\begin{figure}[H]
\centering
\includegraphics[width=\textwidth]{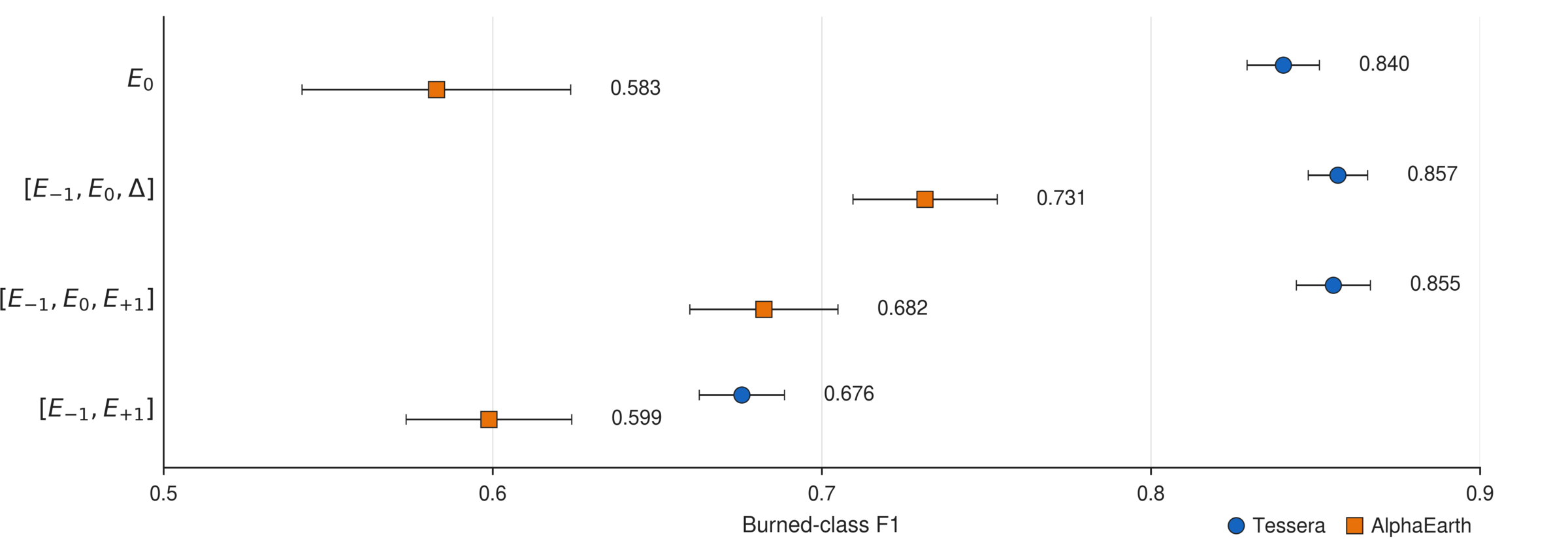}
\caption{Effect of multi-year temporal context on burned-class F1 for Tessera and AlphaEarth. Points show means across three spatial cross-validation folds; error bars indicate $\pm$1 standard deviation. $E_{-1}$, $E_0$, and $E_{+1}$ denote the years before, containing, and after the fire, respectively, and $\Delta=E_0-E_{-1}$. Full details are provided in Section~\ref{supp:multiyear}.}
\label{fig:temporal-context}
\end{figure}

The two embedding models also responded differently when the temporal input was expanded beyond the fire year (Figure~\ref{fig:temporal-context}). In a three-fold Random Forest ablation, using all three years as a concatenated input, $[E_{-1},E_0,E_{+1}]$, improved AlphaEarth F1 from 0.583 to 0.682, while leaving Tessera essentially unchanged (0.840 to 0.855). Providing the pre-fire and fire-year embeddings together with their explicit difference, $[E_{-1},E_0,E_0-E_{-1}]$, produced the strongest AlphaEarth result (F1 = 0.731), while again providing only a modest improvement for Tessera (F1 = 0.857). Interestingly, using only the pre- and post-fire years, $[E_{-1},E_{+1}]$, performed similarly to $E_0$ alone for AlphaEarth (F1 = 0.599 versus 0.583), whereas for Tessera it substantially reduced performance (0.676 versus 0.840). This emphasizes the importance of $E_0$ for Tessera and the greater benefit of cross-year context for AlphaEarth. Together with the latent-space analysis, these results are consistent with Tessera encoding the disturbance itself prominently within the annual representation that contains the event, whereas AlphaEarth appears to benefit more from information distributed across annual representations. Complete results are provided in Section~\ref{supp:multiyear} of the Supplementary Material.

The source of this difference remains unclear, as the models differ in training data, input sensors, spatial context, architecture, loss functions, temporal sampling, and training targets. One notable pattern is that adding spatial context through U-Net models improved performance much more for AlphaEarth than for Tessera, despite AlphaEarth embeddings already containing patch-level spatial context. The pattern suggests that when a pixel's annual trajectory already contains a clear disturbance signal, strong temporal encoding reduces dependence on spatial decoding. AlphaEarth's latent projections also seem more strongly organized by land cover in the fire year, which may partly reflect differences in training objectives, including its use of land-cover information as a target. Without systematic ablations, however, we cannot attribute the observed differences to any single design choice. Future work should isolate the effects of spatial context, temporal sampling, input modalities, training targets, and loss functions to determine what makes annual embeddings more sensitive to abrupt disturbance.

\subsection{Annual embeddings match paired pre- and post-fire imagery for burned-area delineation}

Annual embeddings substantially simplify burned area mapping workflows by moving data preparation and temporal reasoning upstream into a reusable representation. Precomputed embeddings can be downloaded directly for a region and year of interest, without requiring the downstream user to perform cloud masking, image filtering, dense time-series processing, or task-specific feature construction. The experiments in this study showed that, once this representation is available, a single fire-year embedding paired with simple downstream models can map burned area as accurately as paired pre- and post-fire imagery, while also retaining enough temporal information to recover the approximate date of the event.

The contrast is clearest against event-specific burned area delineation. These approaches require carefully selected cloud-free, well-timed post-fire imagery, and sometimes paired pre-fire imagery. This selection is not trivial: it requires knowing where and approximately when a fire occurred, and it must balance cloud contamination, image quality, and vegetation-dependent recovery timing. An image acquired too long after a grassland or shrubland fire may show a landscape that has already recovered, while one acquired too early may capture a fire still in progress, as for the Lions Fire in Figure~\ref{fig:singlefire_maps}. Poor segmentation performance can then reflect either an inadequate model or a poorly chosen input, and the two sources of error are difficult to separate. Crucially, image selection is required not only for training but also at inference, so these models cannot be applied to an arbitrary landscape unless suitable post-fire images have first been identified. This limits the scalability of event-specific workflows. An embedding workflow uses the same input everywhere: the precomputed annual embedding.

This simplicity comes at the cost of latency. Current annual embedding products are generated only after the annual observation period is complete, so they cannot support within-season or near-real-time burned area mapping in the way that online time-series or event-specific imagery workflows can. Their current strength is therefore retrospective, large-scale mapping rather than rapid fire monitoring.

\citet{silvaEvaluatingAlphaEarthTESSERA2026} found a different ranking in Portugal, where a pre- and post-fire Sentinel-2 baseline outperformed pixel-level classifiers on Tessera embeddings (F1 = 0.895 vs.\ 0.860). Our two studies differ in scale and evaluation design: their study used 20 fires and evaluated point samples drawn from a ring immediately surrounding each perimeter, whereas ours used 559 single-fire chips and evaluated every pixel across 15~km chips. Their results also illustrate the single-fire reference problem that motivated our protocol: in their largest mapped example, most of the apparent false positives coincided with another 2024 fire. While the relative ranking may depend on dataset scale and evaluation design, the two studies agree on the essential point: annual Tessera embeddings approach or match fire-specific pre/post workflows without event-specific image construction.

\subsection{Annual embeddings support regional mapping, but precision limits wall-to-wall deployment}

Annual embeddings also simplify the operational problem of mapping all burned area within a year. Dense time-series approaches traverse many observations to identify candidate events, compare pre- and post-fire conditions, and merge duplicate detections across consecutive images that all carry the burn signal. Annual embeddings collapse this temporal search into a single representation for the summary period. The temporal reasoning is not eliminated, but shifted upstream into the foundation model and made reusable across downstream tasks. In the benchmark, the Tessera Lightweight U-Net maintained almost the same accuracy when the target expanded from the single benchmark fire to all same-year burned area within each chip (F1 = 0.911 vs.\ 0.896), while recovering additional fires absent from the original single-fire labels.

The California experiment shows why this matters at medium spatial resolution. A model trained entirely outside the state detected substantially more small and medium-sized fires than GABAM or MCD64A1, while maintaining high spatial agreement for larger events. More strikingly, manual inspection of the largest unmatched predictions identified genuine fires absent from both MTBS and CAL FIRE, despite California having unusually comprehensive fire inventories. This reinforces the motivation for medium-resolution regional products: finer mapping improves the delineation of known fire perimeters and can also recover smaller events missed by coarser products or incomplete inventories.

Wall-to-wall deployment nevertheless exposes a challenge that is largely hidden in fire-centred benchmarks: the model must distinguish wildfire from the full range of non-fire trajectories encountered across a landscape. Mature burned area products devote substantial processing to suppressing commission errors through temporal, spectral, land-cover, and spatial constraints. The wall-to-wall inference in this study applied no comparable precision-oriented filtering: the model was evaluated out of the box, with no statewide negative sampling and minimal post-processing. Although statewide pixel-level precision remained comparable to MCD64A1, Tessera produced substantially more unmatched components. The comparison should therefore not be read as a benchmark between equivalent products, but between mature pipelines with explicit mechanisms for suppressing commission errors and a lightweight U-Net trained on fire-centred chips outside California.

That such a minimal pipeline matched a mature product on statewide pixel-level precision while detecting far more small fires is the encouraging half of the result: Tessera embeddings encode a transferable burned area signal. The remaining challenge is controlling the larger number of unmatched detections without sacrificing that small-fire sensitivity, and the structure of the errors suggests this may be tractable. Confirmed false positives were concentrated in snow-covered high-elevation terrain in the southern Sierra Nevada, the hills surrounding Los Angeles, and agricultural regions, consistent with non-fire trajectories that were poorly represented in fire-centred training data. Errors of this kind appear amenable to targeted improvements rather than requiring a different representation altogether.

\subsection{Annual embeddings retain approximate intra-annual event timing}

The timing experiment shows that annual Tessera embeddings retain recoverable information about when a fire occurred within the year, even though the embedding is a single annual representation rather than an explicit time series. This is likely a consequence of how Tessera summarizes temporal observations: every Sentinel-1 and Sentinel-2 input is associated with its day of year during both training and inference, so the model can encode not only that a trajectory was disturbed, but where within the annual sequence the disturbance occurred. A pixel that burns in March, July, or November follows a different yearly trajectory even if all three end the year in a burned state, and the timing results indicate that some of this temporal ordering survives in the annual summary. A US-trained model dated European fires to within 10.5 days on average, which suggests that the encoded timing is not specific to US fire regimes.

Event dating matters operationally: burned area products are expected to report not only where fires occurred but also when. Timing supports analysis of seasonal and climatic fire patterns and evaluation of policies that restrict burning during periods of elevated fire risk. Many existing burned area products derive timing from dense observation sequences, whereas annual composites provide much coarser temporal information. Annual embeddings sit between these approaches: they do not preserve the full observation sequence, yet they retain substantially more timing information than a conventional annual composite.

Event dating is tightly coupled to disturbance detection, as it is in any remote sensing approach. A dense time-series method that never registers a burn signal cannot identify the first burned observation, and an embedding-based pipeline that fails to detect a burned area has no event to date. For annual embeddings, the additional constraint is that the relevant temporal information must have been preserved during representation learning. This is visible for late-season fires, where burned area recall declines and predicted day of year is underestimated for fires that remain detected. Timing from annual embeddings should therefore be interpreted as conditional on the disturbance being both detected and sufficiently encoded in the annual representation.

A final caveat concerns the reference labels. The ground truth assigns a single ignition day of year to every pixel within a fire perimeter, even though large fires spread over days or weeks; the model was trained on these approximate pixel-level labels and evaluated after averaging pixel predictions to the fire level. This is a practical first test rather than true per-pixel burn dating, and future work should distinguish ignition date, pixel-level burn date, and the full active fire period, possibly through object-level formulations rather than per-pixel regression. The present results should be read as evidence that approximate intra-annual timing is recoverable from annual embeddings, laying the foundation for more detailed timing estimation.

\subsection{Late-season fires and the limits of calendar-year summaries}

Models trained on Tessera embeddings mapped burned area accurately across most of the fire season, but recall declined sharply for fires that ignited late in the calendar year, particularly in November and December. A similar but weaker late-season decline was apparent with AlphaEarth. For both embedding products, the association between later ignition date and lower recall persisted after accounting for fire characteristics and geography. This pattern is therefore consistent with a limitation related to annual temporal summarization, although its effect was more pronounced for Tessera.

The practical impact of this limitation depends on geography. In California, relatively few fires occur in the final weeks of the calendar year, so only a small share are affected. But in regions where the fire season may span the calendar-year boundary, such as southern Australia \citep{williamsFlammableAustraliaFire2012}, a substantial share of fires could fall in the poorly encoded end-of-year window. A 12-month summary is a sensible default, but its alignment to the calendar year is arbitrary, and it aligns with some fire regimes far better than others.

Why do annual embeddings encode late-season fires poorly? The most direct explanation is limited observational support: a November fire leaves only weeks of post-fire observations within the summary window, and our results confirmed that accuracy degrades severely when no cloud-free post-fire imagery is available. Observation scarcity does not explain every failure, however. Several late-season fires had cloud-free post-fire images and were still poorly detected, suggesting that a short post-fire period may produce a disturbance signature too weak to shape the annual summary. Inference-time sampling may compound this: Tessera v1 randomly samples 40 observations per pixel, so the few post-fire observations available after a late-season fire may be sparsely sampled or missed entirely. This sampling strategy has changed in later Tessera versions \citep{fengTESSERAV2Scaling2026a}, which may improve late-season encoding. Finally, the pre-training data may under-represent late-season fires in US ecosystems, biasing the model toward mid-year disturbance; however, this cannot be assessed from the present experiments. Future work should establish whether the late-season failure is specific to US fire regimes, how it behaves in Southern Hemisphere fire seasons, and whether it can be mitigated by shifting the summary window or by pre-training on data with greater temporal diversity and more examples of change and disturbance.

\section{Conclusions}

This study asked whether annual Earth-observation embeddings encode wildfire disturbance well enough to map burned area without curated post-fire imagery or dense time-series processing. For Tessera, the answer to each of the questions posed in the Introduction is largely yes. Wildfire disturbance is a dominant, directly readable feature of the fire-year embedding; a single embedding matched paired pre- and post-fire imagery for delineating individual burn scars; the same models mapped all same-year fires within a landscape, an unseen region, and another continent without retraining; and the embedding retained enough temporal structure to date detected fires to within about two weeks. AlphaEarth encodes the same disturbance more weakly, and extracts it best when spatial context or explicit temporal contrast is supplied downstream.

Embeddings therefore offer elements of both families of approaches: they can approach the delineation accuracy of event-specific methods without curated post-fire imagery, while supporting wall-to-wall mapping without dense temporal processing at the downstream stage. Additional complexity can still be layered on through spatial decoders, multi-year inputs, calibration, or post-processing, but even the simple pipeline evaluated here produces competitive burned area maps.

Several limitations remain. Fires that ignite in the last weeks of the calendar year are poorly encoded by annual summaries aligned to the calendar year, and current annual embedding products are inherently retrospective: the full representation is not available until the annual observation period is complete. Wall-to-wall application also produced substantially more suspected false positives than mature burned area products, although these represented only a small share of predicted burned area and manual inspection showed that some were genuine fires absent from the reference datasets. Operational deployment would require additional calibration, more diverse negative examples, and post-processing to distinguish wildfire from other fire-like trajectories; training on unburned areas that experienced other kinds of change, which recent work has shown to substantially improve precision \citep{liuFasterBetterMore2026}, is particularly promising. Stable embedding products and clear versioning will also be important, since new releases may require downstream models to be retrained or recalibrated. Whether the disturbance sensitivity shown here for wildfire extends to disturbance more generally remains an open question for future work.

\section*{Data availability}

The source datasets used in this study are publicly available from the sources cited in the manuscript. The derived data and code supporting the findings can be made available upon request.

\section*{Declaration of competing interest}

Clement Atzberger is affiliated with dClimate, a commercial company that has independently invested resources in generating a global 2017--2025 archive of Tessera v1.1 embeddings and making it available through AWS Open Data. The present study used Tessera v1 embeddings and did not use the v1.1 archive. The remaining authors declare no competing financial interests or personal relationships that could have appeared to influence the work reported in this paper.

\section*{Acknowledgments}

We thank Samuel Barrett, Anil Madhavapeddy, Aneesh Naik and Sadiq Jaffer for their careful reading of earlier versions of the manuscript and their constructive comments and discussions.

This work was made possible by openly available Earth-observation data and products. We gratefully acknowledge the European Space Agency and the Copernicus Programme for Sentinel-1 and Sentinel-2 data; NASA for Harmonized Landsat--Sentinel and MODIS data; the USDA Forest Service and US Geological Survey for Monitoring Trends in Burn Severity data; CAL FIRE for the Fire and Resource Assessment Program fire-perimeter database; and the Copernicus Emergency Management Service for the Rapid Mapping products. We also acknowledge the teams responsible for GABAM, LCMAP, RESOLVE Ecoregions, Tessera, and AlphaEarth Foundations; the GeoTessera developers and the Taylor Geospatial Institute and Source Cooperative for facilitating access to the annual embeddings; and Google Earth Engine for providing access to data and computational resources.

We acknowledge support from the Tezos Foundation and the UKRI-funded AI4ER doctoral training programme.

\section*{CRediT authorship contribution statement}

\textbf{Jovana Knezevic:} Conceptualization, Methodology, Software, Data curation, Formal analysis, Investigation, Validation, Visualization, Project administration, Writing -- original draft, Writing -- review \& editing. \textbf{Clement Atzberger:} Conceptualization, Methodology, Writing -- review \& editing. \textbf{Zhengpeng Feng:} Methodology, Writing -- review \& editing. \textbf{Adam F. A. Pellegrini:} Methodology. \textbf{Srinivasan Keshav:} Conceptualization, Supervision, Writing -- review \& editing, Resources. \textbf{David Coomes:} Conceptualization, Supervision, Writing -- review \& editing, Resources.

\section*{Declaration of generative AI and AI-assisted technologies in the writing process}

During the preparation of this work, we used ChatGPT (OpenAI) and Claude models (Anthropic) accessed through Cursor for coding assistance, and for editing and providing feedback on author-drafted text. All outputs were reviewed, validated and revised as needed, and the authors take full responsibility for the content of the publication.

\ifdefined\buildcombined\else
\bibliographystyle{elsarticle-harv}
\bibliography{references,manual}
\fi

\ifdefined\buildcombined
\clearpage
\section*{Supplementary Material}

\textit{Annual Earth-observation embeddings encode wildfire disturbance and support simplified burned area mapping}

\renewcommand{\thesection}{S\arabic{section}}
\renewcommand{\thesubsection}{S\arabic{section}.\arabic{subsection}}
\renewcommand{\thefigure}{S\arabic{figure}}
\renewcommand{\thetable}{S\arabic{table}}
\renewcommand{\theequation}{S\arabic{equation}}
\setcounter{section}{0}
\setcounter{figure}{0}
\setcounter{table}{0}
\setcounter{equation}{0}

\section{Paired pre- and post-fire HLS baseline}
\label{supp:prefire-hls}

\subsection{Pre-fire HLS composite construction}

We generated pre-fire imagery from HLS Sentinel-2 (S30) v2.0 surface reflectance (NASA/HLS/HLSS30/v002). Only S30 observations were used. For each chip, the MTBS ignition date (\texttt{ig\_date}) associated with the target fire was used to define the pre-fire period. We initially queried observations within the interval \([t_{\mathrm{ign}}-30\,\mathrm{days},\,t_{\mathrm{ign}})\), thereby excluding observations acquired on or after the ignition date. If valid coverage across the full chip remained below 95\%, the search window was extended to 60 days before ignition.

We retained scenes intersecting the chip geometry and with granule-level \texttt{CLOUD\_COVERAGE}~$<$\,50\%. At the pixel level, observations were masked when any of HLS Fmask bits 0--4 were set, corresponding to cirrus, cloud, adjacent-to-cloud, cloud shadow, or snow/ice. The six surface-reflectance bands used were Blue, Green, Red, NIR, SWIR1, and SWIR2, corresponding to S30 bands B2, B3, B4, B8A, B11, and B12. HLS surface-reflectance values were converted to floating point using the scale factor 0.0001 and were not clipped to the \([0,1]\) interval. Masked values were retained as missing (NaN).

The remaining observations were combined using a temporally weighted medoid. For each pixel, we first calculated the temporal median independently for each of the six bands. For observation $i$, spectral distance from this median was then calculated as
\begin{equation}
d_i=\sum_{b=1}^{6}(x_{i,b}-\tilde{x}_b)^2 ,
\end{equation}
where $x_{i,b}$ is the reflectance of band $b$ in observation $i$, and $\tilde{x}_b$ is its temporal median. Observations were additionally weighted according to their temporal distance from ignition,
\begin{equation}
w_i=\frac{1}{\max(\Delta t_i,1)},
\end{equation}
where $\Delta t_i$ is the number of days between the observation and ignition. The quality score used for medoid selection was proportional to
\begin{equation}
q_i=\frac{1}{d_i/w_i},
\end{equation}
thereby favouring observations that were both spectrally representative of the pre-fire period and closer to the ignition date. The highest-quality observation was selected independently for each pixel, with all six bands retained from the same acquisition. Where a medoid observation could not be obtained, we used the per-band temporal median as a fallback.

For chips that still contained less than 95\% valid coverage after the 60-day search, we repeated the 60-day compositing procedure without the granule-level cloud-coverage filter while retaining the per-pixel Fmask criteria. Chips that already satisfied the coverage criterion were not regenerated.

Pre-fire composites were exported as 512\,$\times$\,512 pixel GeoTIFFs at 30\,m resolution in the UTM projection and spatial footprint of the corresponding benchmark chip. The composite construction was implemented in \texttt{export\_chip\_prefire\_hls\_composite.py}.

\subsection{Post-fire imagery and HLS versions}

Post-fire imagery was taken directly from the HLS Burn Scars benchmark and corresponds to the benchmark-provided HLS S30 v1.4 merged imagery. We retained these images rather than reconstructing the post-fire observations from HLS v2.0 in order to preserve the event-specific imagery and acquisition dates used in the original benchmark.

Equivalent pre-fire imagery is not provided by the benchmark, and the pre-fire composites described above were therefore generated separately from HLS S30 v2.0. Consequently, the paired baseline combines v2.0 pre-fire imagery with v1.4 post-fire benchmark imagery, with the same version pairing used throughout training and evaluation.

Both pre- and post-fire source chips contained six 30\,m surface-reflectance bands over the same 512\,$\times$\,512 pixel footprint. Prior to model input construction, both were aligned to the common PANGAEA 10\,m grid, producing arrays of 1536\,$\times$\,1536\,$\times$\,6 pixels.

\subsection{Paired model input}

For the \texttt{hls\_pre\_post\_diff} configuration used in the main comparison, we constructed a 21-channel input consisting of:
\begin{itemize}
\item six post-fire reflectance bands;
\item six pre-fire reflectance bands;
\item post-fire NBR and pre-fire NBR;
\item differenced NBR,
\begin{equation}
\mathrm{dNBR}=\mathrm{NBR}_{\mathrm{pre}}-\mathrm{NBR}_{\mathrm{post}};
\end{equation}
\item six band-wise reflectance differences,
\begin{equation}
\Delta B=B_{\mathrm{pre}}-B_{\mathrm{post}}.
\end{equation}
\end{itemize}
Thus, the paired baseline supplied the downstream model with both the two fire-specific observations and explicit spectral-change features. The post-fire-only HLS baseline used only the six benchmark post-fire reflectance bands.

\subsection{Input validity and missing pixels}

Chips were retained for the paired comparison when both pre- and post-fire inputs contained more than 80\% valid pixels. Of the 577 eligible single-fire chips, 18 failed this criterion, leaving 559 chips for the paired comparison. Missing coverage among retained chips was generally negligible: median invalid coverage was 0\% for both inputs, while mean invalid fractions were 0.29\% for the pre-fire composites and 0.16\% for the post-fire imagery. Only 9 of 559 retained chips contained more than 10\% invalid coverage in either input.

Pixels containing non-finite input features were excluded when constructing training samples. During full-chip inference, pixels for which the required input features were invalid were assigned an unburned prediction. Pixels marked as nodata in the MTBS reference were excluded from accuracy evaluation.
\section{Additional methods and label refinement details}
\label{supp:labels}

\begin{figure}[H]
\centering
\includegraphics[width=\textwidth]{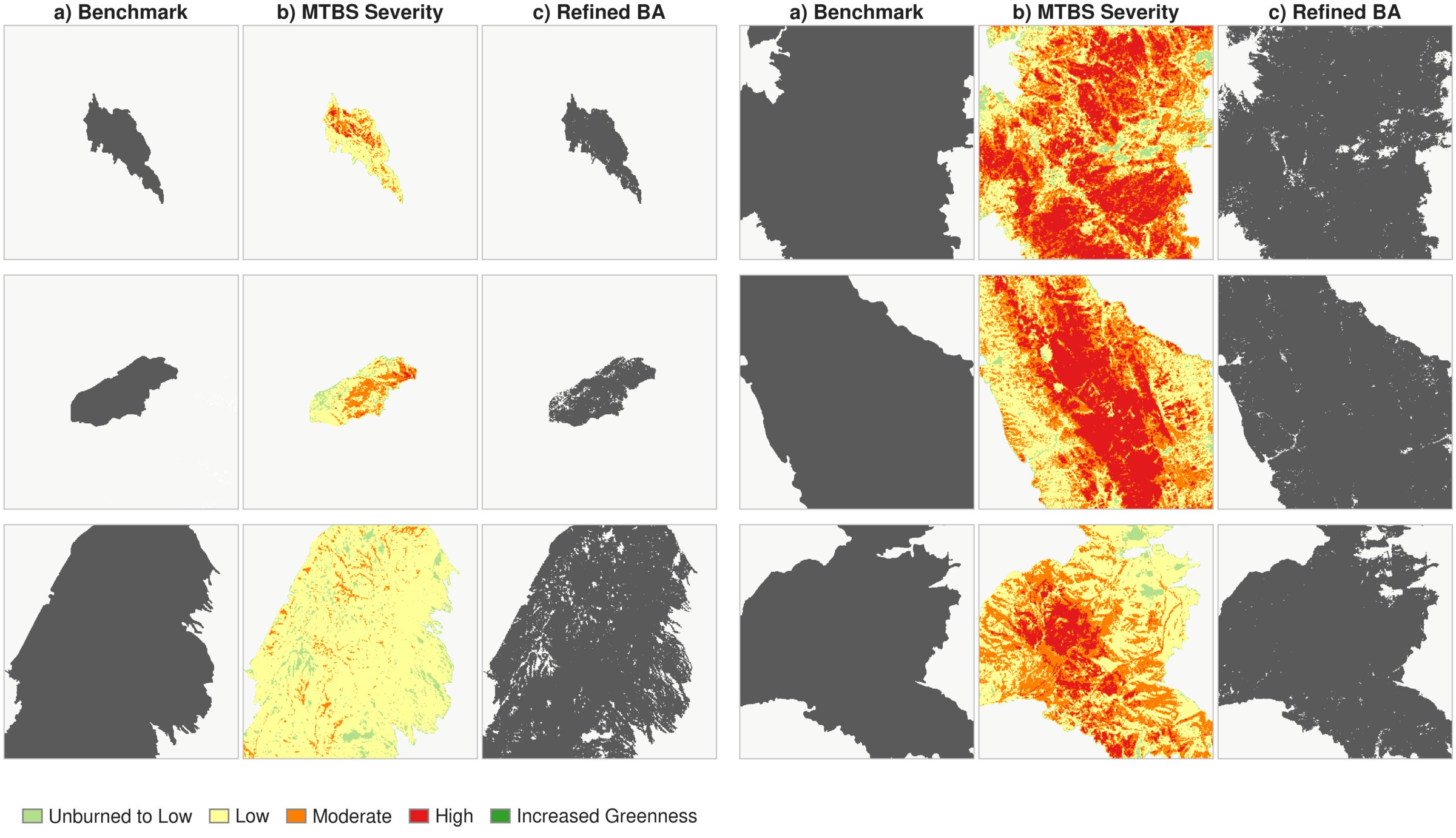}
\caption{Examples of HLS Burn Scars benchmark labels and their refinement using MTBS burn-severity data. For each example, columns show the original benchmark burned area mask, the corresponding MTBS burn-severity mosaic, and the refined binary burned area mask used in this study. MTBS severity classes 2 (low), 3 (moderate), and 4 (high) were treated as burned, while classes 0 (unburned), 1 (unburned to low), and 5 (increased greenness) were treated as unburned. The examples illustrate how severity-based refinement removes unburned and spectrally ambiguous areas contained within the original benchmark masks.}
\label{fig:supp_label_refinement}
\end{figure}
\section{Downstream model implementation}
\label{supp:downstream-models}

The downstream models used throughout the burned area mapping experiments comprised L2-regularized logistic regression, Random Forest (RF), a standard U-Net, and a lightweight U-Net. The main model configurations are summarized in Table~\ref{tab:downstream_models} of the main text; additional implementation details are provided here.

\paragraph{Logistic regression.}
Logistic regression was implemented using scikit-learn. Input features were standardized using \texttt{StandardScaler}, fitted on the training data and applied to the corresponding validation and test data. We used L2 regularization with $C=1$ and balanced class weights.

\paragraph{Random Forest.}
RF was implemented using scikit-learn with 200 trees and no class weighting. At each split, $\sqrt{d}$ of the $d$ input features were considered. Tree depth was unrestricted, with a minimum of five samples per leaf and ten samples required to split an internal node.

\paragraph{Standard U-Net.}
The standard U-Net was implemented using the \texttt{segmentation\_models\_pytorch} library with a ResNet-18 encoder initialized from random weights, without ImageNet pretraining. The decoder contained five stages with 256, 128, 64, 32, and 16 channels, respectively. The resulting model contained approximately 14.5 million trainable parameters.

The model was trained using AdamW with an initial learning rate of $10^{-4}$, weight decay of $10^{-4}$, and cosine learning-rate annealing to $10^{-6}$. The loss function was the sum of binary cross-entropy and Dice loss. Training used mixed precision and gradient clipping with a maximum norm of 1.0. Models were trained with a batch size of 32 for up to 50 epochs, with early stopping after 10 epochs without improvement in validation loss. Training samples consisted of $256 \times 256$ pixel crops, with eight crops randomly sampled from each training chip per epoch. Random horizontal and vertical flips were used for data augmentation.

\paragraph{Lightweight U-Net.}
The lightweight U-Net was implemented as a custom three-level encoder-decoder with skip connections and a base channel width of 32. Each convolutional block used $3\times3$ convolutions followed by batch normalization and ReLU activation. Spatial downsampling was performed using max pooling, while the decoder used bilinear upsampling followed by concatenation with the corresponding encoder features. A $1\times1$ convolutional classification head produced the final per-pixel output. The model contained approximately 0.87 million trainable parameters.

The lightweight U-Net used the same loss function, optimizer, learning-rate schedule, weight decay, augmentation, mixed-precision training, gradient clipping, crop-sampling strategy, and early-stopping criterion as the standard U-Net. The only difference in the training configuration was a larger batch size of 64.
\section{Additional 2D projections}
\label{supp:umap}

\begin{figure}[H]
\centering
\includegraphics[width=\textwidth]{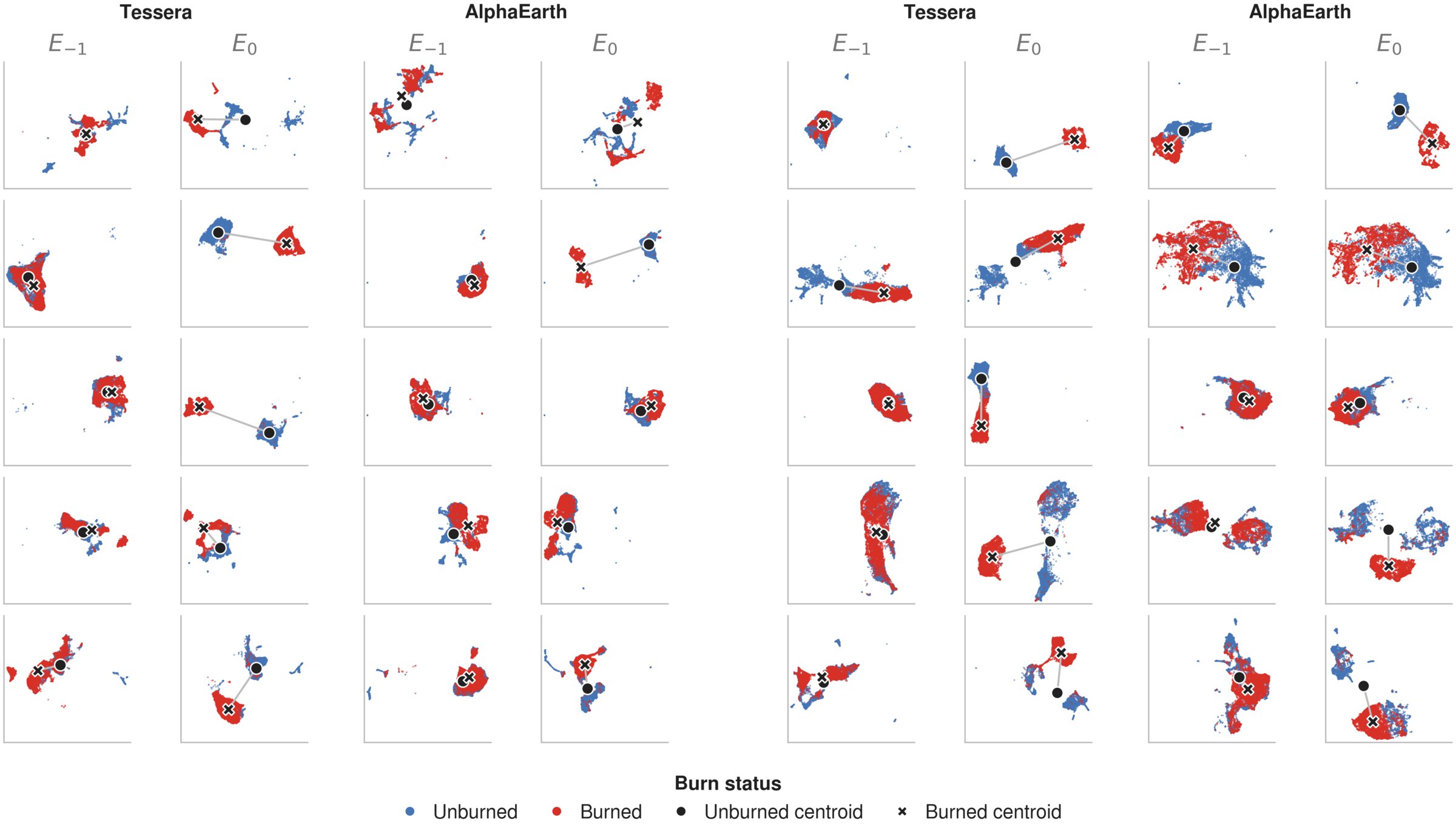}
\caption{Nonlinear UMAP projections for a sample of 10 randomly selected chips. For each embedding product, UMAP was fitted jointly to the pre-fire ($E_{-1}$) and fire-year ($E_0$) embeddings. Points are coloured by burn status; markers show the burned and unburned centroids.}
\label{fig:supp_umap}
\end{figure}

\begin{figure}[H]
\centering
\includegraphics[width=\textwidth]{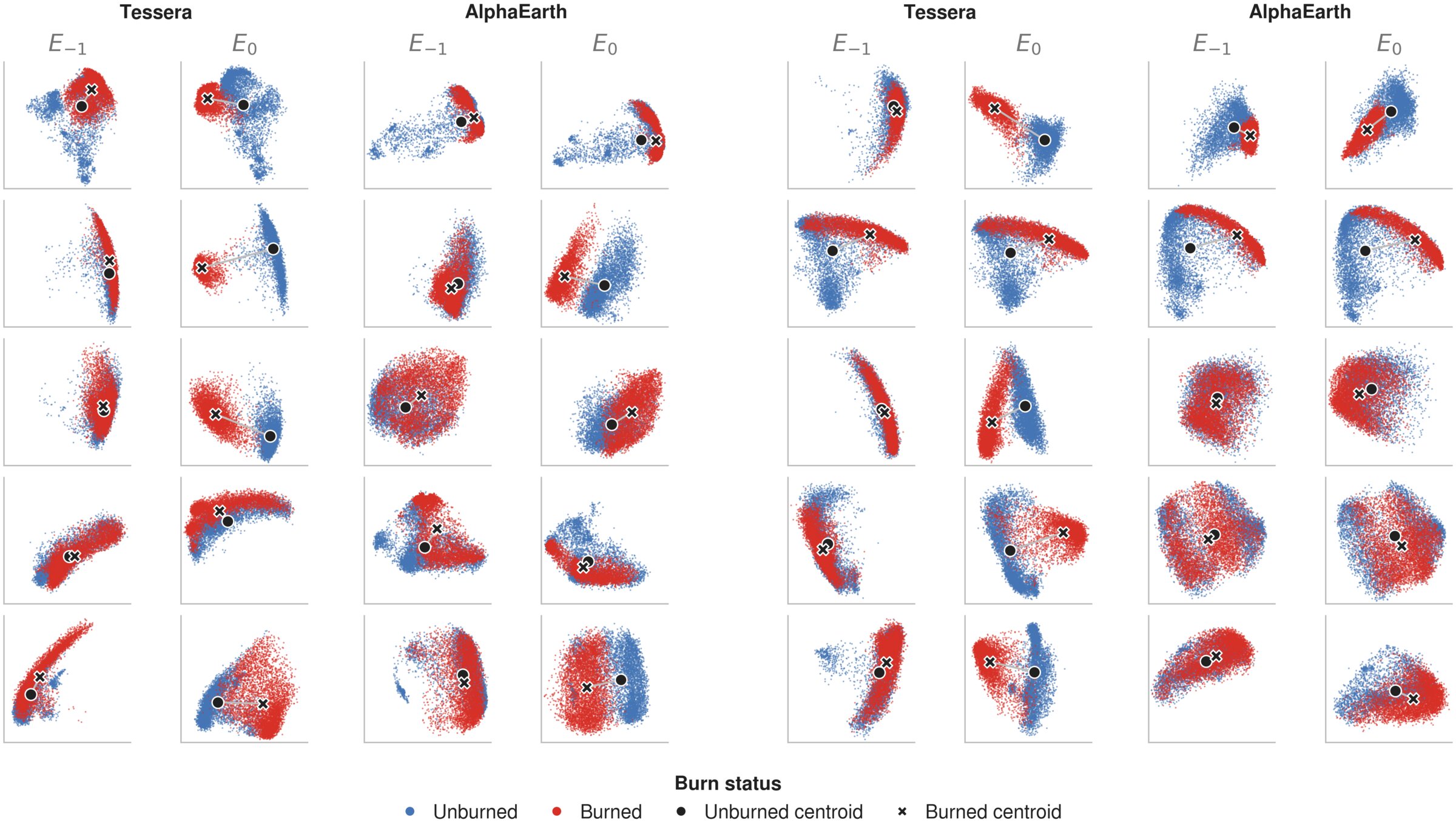}
\caption{PCA projections for a sample of 10 randomly selected chips. For each embedding product, PCA was fitted jointly to the pre-fire ($E_{-1}$) and fire-year ($E_0$) embeddings. Points are coloured by burn status; markers show the burned and unburned centroids.}
\label{fig:supp_pca}
\end{figure}
\clearpage
\section{Single-fire delineation metrics}
\label{supp:single-fire-metrics}

\begin{table}[p]
\centering
\caption{Single-fire burned area delineation on 559 chips with more than 80\% valid pre- and post-fire HLS coverage, evaluated with spatially blocked three-fold cross-validation. Metrics are burned-class scores from full-chip pooled evaluation (fold mean~$\pm$~std). MTBS severity classes 1 and 5 were counted as unburned. Tessera, AlphaEarth, and post-fire HLS results were re-evaluated on this cohort; the pre- and post-fire HLS baseline was trained on it originally.}
\label{tab:supp_singlefire_metrics}
\footnotesize
\setlength{\tabcolsep}{3pt}
\renewcommand{\arraystretch}{1.05}
\resizebox{\textwidth}{!}{%
\begin{tabular}{@{}
>{\raggedright\arraybackslash}p{1.5cm}
>{\raggedright\arraybackslash}p{1.25cm}
c
*{5}{r}
@{}}
\toprule
Input & Model & Folds & Accuracy & F1 & IoU & Precision & Recall \\
\midrule
\multirow[t]{4}{*}{Tessera} & LogReg & 3 & $0.947 \pm 0.003$ & $0.791 \pm 0.013$ & $0.655 \pm 0.017$ & $0.704 \pm 0.016$ & $0.903 \pm 0.015$ \\
 & RF & 3 & $0.971 \pm 0.002$ & $0.870 \pm 0.014$ & $0.771 \pm 0.023$ & $0.857 \pm 0.013$ & $0.884 \pm 0.019$ \\
 & Light U-Net & 3 & $0.980 \pm 0.002$ & $0.911 \pm 0.011$ & $0.837 \pm 0.019$ & $0.897 \pm 0.014$ & $0.927 \pm 0.010$ \\
 & U-Net & 3 & $0.981 \pm 0.001$ & $0.916 \pm 0.008$ & $0.845 \pm 0.014$ & $0.903 \pm 0.012$ & $0.930 \pm 0.014$ \\
\addlinespace[2pt]
\multirow[t]{4}{*}{AlphaEarth} & LogReg & 3 & $0.728 \pm 0.021$ & $0.386 \pm 0.025$ & $0.239 \pm 0.020$ & $0.258 \pm 0.021$ & $0.769 \pm 0.017$ \\
 & RF & 3 & $0.892 \pm 0.004$ & $0.585 \pm 0.013$ & $0.414 \pm 0.013$ & $0.508 \pm 0.012$ & $0.691 \pm 0.014$ \\
 & Light U-Net & 3 & $0.949 \pm 0.002$ & $0.765 \pm 0.019$ & $0.620 \pm 0.025$ & $0.778 \pm 0.014$ & $0.753 \pm 0.024$ \\
 & U-Net & 3 & $0.960 \pm 0.003$ & $0.806 \pm 0.021$ & $0.676 \pm 0.030$ & $0.872 \pm 0.027$ & $0.750 \pm 0.023$ \\
\addlinespace[2pt]
\multirow[t]{4}{*}{\makecell[tl]{Post-fire\\HLS}} & LogReg & 3 & $0.897 \pm 0.004$ & $0.650 \pm 0.016$ & $0.482 \pm 0.018$ & $0.520 \pm 0.015$ & $0.868 \pm 0.022$ \\
 & RF & 3 & $0.910 \pm 0.006$ & $0.679 \pm 0.011$ & $0.514 \pm 0.012$ & $0.560 \pm 0.015$ & $0.862 \pm 0.026$ \\
 & Light U-Net & 3 & $0.967 \pm 0.003$ & $0.856 \pm 0.010$ & $0.748 \pm 0.016$ & $0.837 \pm 0.019$ & $0.875 \pm 0.016$ \\
 & U-Net & 3 & $0.969 \pm 0.002$ & $0.860 \pm 0.011$ & $0.755 \pm 0.017$ & $0.850 \pm 0.012$ & $0.871 \pm 0.025$ \\
\addlinespace[2pt]
\multirow[t]{4}{*}{\makecell[tl]{Pre- and\\post-fire HLS}} & LogReg & 3 & $0.947 \pm 0.002$ & $0.791 \pm 0.011$ & $0.654 \pm 0.015$ & $0.701 \pm 0.013$ & $0.906 \pm 0.011$ \\
 & RF & 3 & $0.949 \pm 0.006$ & $0.798 \pm 0.024$ & $0.665 \pm 0.033$ & $0.706 \pm 0.032$ & $0.920 \pm 0.010$ \\
 & Light U-Net & 3 & $0.976 \pm 0.004$ & $0.896 \pm 0.015$ & $0.812 \pm 0.025$ & $0.872 \pm 0.017$ & $0.921 \pm 0.017$ \\
 & U-Net & 3 & $0.979 \pm 0.003$ & $0.905 \pm 0.011$ & $0.827 \pm 0.019$ & $0.890 \pm 0.020$ & $0.921 \pm 0.007$ \\
\bottomrule
\end{tabular}%
}

\vspace{0.5em}
{\footnotesize RF = Random Forest; LogReg = Logistic Regression.}
\end{table}
\section{Late-season fires analysis}
\label{supp:late-season}

To assess whether the decline in recall for fires igniting late in the calendar year could be explained by differences in fire characteristics or geography, we fitted separate quasi-binomial generalized additive models for each embedding product and downstream model. The analysis used one observation per unique MTBS fire event, with each fire weighted equally. Fires intersecting overlapping chips were excluded because detected pixels could not be assigned unambiguously, leaving 1{,}031 fires across 51 ecoregions. Recall was modelled as a penalized smooth of the number of days remaining in the calendar year after ignition, with an additional smooth for log fire area and adjustment for the proportion of moderate- or high-severity burned area, incident type, dominant pre-fire land cover, biome, and cross-validation fold. Adjusted recall curves were obtained by predicting each fire across the timing range while retaining its observed values for all other covariates, and then averaging these predictions across fires. Confidence intervals were estimated from 1{,}000 bootstrap resamples clustered by ecoregion. To avoid poorly supported estimates at the calendar-year boundaries, fitted curves and contrasts were restricted to the 2nd--98th percentile of observed ignition timing.

\subsection{Results}

Fire-level recall declined sharply for fires ignited late in the calendar year, for both Tessera and AlphaEarth. Adjustment for fire characteristics, land cover, biome, and cross-validation fold changed this relationship only modestly (Figure~\ref{fig:supp_late_season}). For Tessera, adjusted recall was 0.51 at 60 days remaining in the year and 0.84 at 180 days for the Random Forest, a difference of $-0.33$ (95\% CI: $-0.42$ to $-0.26$); for the Light U-Net the corresponding estimates were 0.64 and 0.85, a difference of $-0.21$ ($-0.33$ to $-0.15$). AlphaEarth recall was lower throughout the year, with mid-year values of 0.69 and 0.64 rather than 0.84 and 0.85, and showed declines of comparable size ($-0.21$, $-0.34$ to $-0.10$, for the Random Forest; $-0.27$, $-0.36$ to $-0.15$, for the Light U-Net). Results with the Lightweight U-Net showed the same late-season limitation for Tessera but stronger model-dependent seasonal variation for AlphaEarth (Figure~\ref{fig:supp_late_season}). The decline also persisted when the analysis was restricted to wildfires in every case, with differences of $-0.14$ to $-0.19$, indicating that it was not explained by the seasonal distribution of prescribed fires. Detection of late-season fires is therefore a shared limitation of the analyzed annual embeddings, and a key consideration for operational use.

\begin{figure}[H]
\centering
\includegraphics[width=\textwidth]{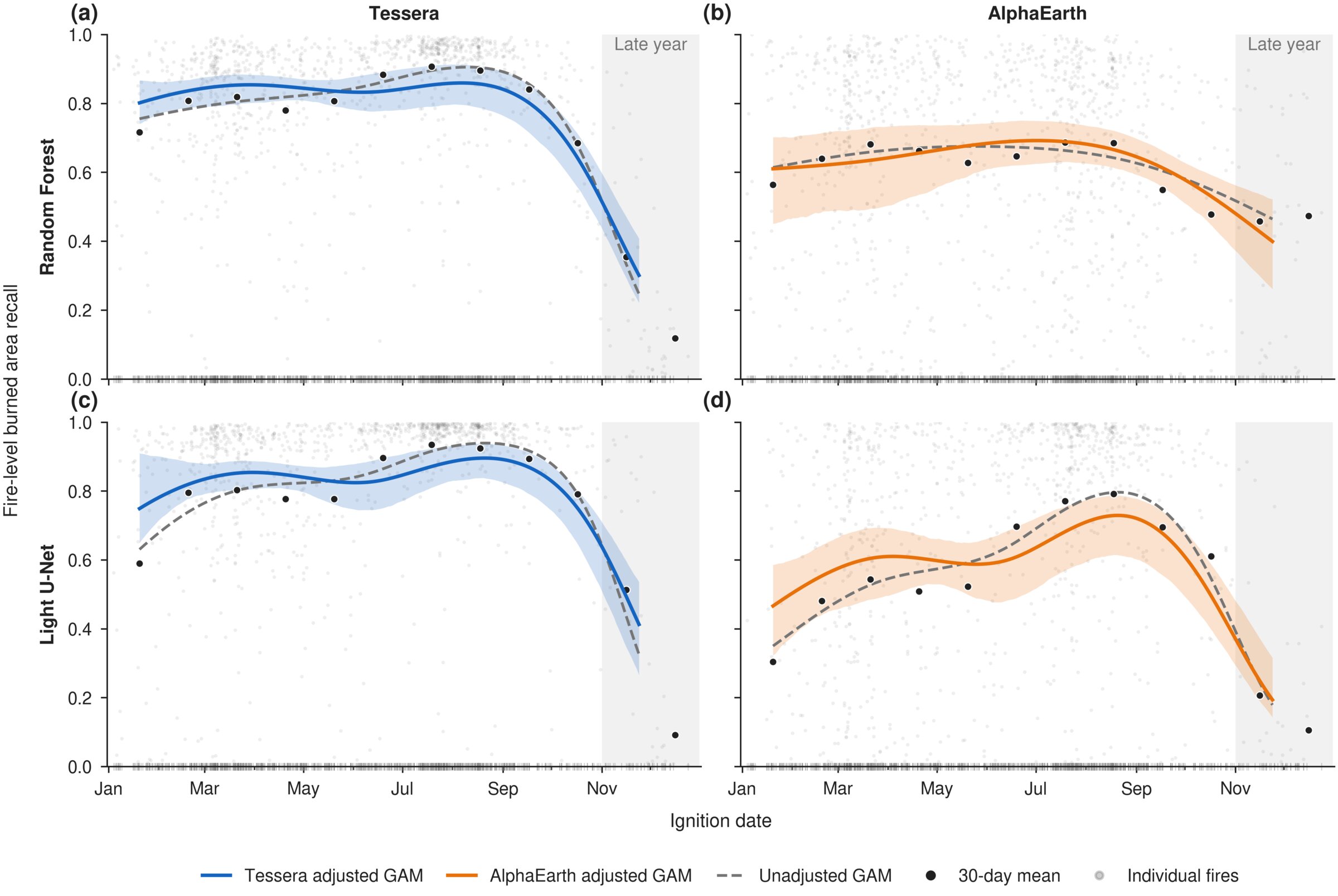}
\caption{Fire-level burned area recall against ignition date: (a)~Tessera and (b)~AlphaEarth with the Random Forest; (c)~Tessera and (d)~AlphaEarth with the Light U-Net. The models were fitted on the same 1{,}031 MTBS fire events. Faint grey points are individual fires and the rug beneath each axis shows their timing distribution; black points are observed mean recall in 30-day bins. The dashed grey line is the unadjusted GAM of recall on ignition timing alone; the coloured line is the adjusted GAM, with a 95\% ecoregion cluster bootstrap interval as the band. Adjusted curves are standardized average marginal predictions: predicted recall averaged over all fires with fire size, severity, incident type, pre-fire land cover, biome and cross-validation fold held at their observed values.}
\label{fig:supp_late_season}
\end{figure}
\section{Contextual comparison with the AlphaEarth transfer results of Seydi}
\label{supp:seydi}

We additionally evaluated Tessera on the 17 European EMSR fires used by Seydi \citep{seydiDeepLearningBasedBurned2025a} to provide a contextual comparison with the published AlphaEarth transfer experiment. Fifteen of these fires are from 2024 and are included in our main 88-fire European evaluation; the remaining two are from 2022 and were downloaded separately for this comparison. This analysis is therefore distinct from the main 2024--2025 transfer experiment.

On the 17 EMSR fires previously evaluated by Seydi \citep{seydiDeepLearningBasedBurned2025a}, the Tessera model exceeded the published AlphaEarth transfer result despite using a simpler input and classifier. Both approaches trained on US fires with MTBS labels and evaluated on the common set of European fires. However, Seydi used a bi-temporal Siamese U-Net over two consecutive years of AlphaEarth embeddings \citep{seydiDeepLearningBasedBurned2025a}, whereas our model used a single fire year Tessera embedding with a 0.87M-parameter lightweight U-Net. Tessera reached F1 = 0.828 $\pm$ 0.180 and IoU = 0.737 $\pm$ 0.195, compared with 0.735 $\pm$ 0.186 and 0.612 $\pm$ 0.220 for AlphaEarth (Table~\ref{tab:spatial_transfer_comparison}). The gain came mainly from recall (0.894 vs 0.749), with precision nearly unchanged (0.824 vs 0.806), indicating that the single-year Tessera embedding recovered more of each burn scar without a comparable loss in precision.

\begin{table}[H]
\centering
\caption{Contextual comparison with Seydi (2025) on a European EMSR subset of 17 fires ($n=17$). Both methods are trained on US fires and evaluated on the same European EMSR subset; metrics are reported as mean $\pm$ standard deviation across fires. Seydi results are taken from the published AlphaEarth transfer experiment.}
\label{tab:spatial_transfer_comparison}
\small
\begin{tabular}{llcccc}
\toprule
Approach & Representation & Precision $\uparrow$ & Recall $\uparrow$ & F1 $\uparrow$ & IoU $\uparrow$ \\
\midrule
Seydi (2025) & AlphaEarth & $0.806 \pm 0.211$ & $0.749 \pm 0.231$ & $0.735 \pm 0.186$ & $0.612 \pm 0.220$ \\
This study & Tessera & $\mathbf{0.824 \pm 0.113}$ & $\mathbf{0.894 \pm 0.210}$ & $\mathbf{0.828 \pm 0.180}$ & $\mathbf{0.737 \pm 0.195}$ \\
\bottomrule
\end{tabular}
\end{table}

The one clear outlier, EMSR638 (F1 = 0.18), is the only November--December fire in the European dataset, matching the late-season encoding limitation identified in the annual fire mapping experiment.
\section{Contextual comparison with PANGAEA results}
\label{supp:pangaea}

\subsection{Methods}

The HLS Burn Scars benchmark was originally created for the evaluation of the Prithvi foundation model \citep{jakubikFoundationModelsGeneralist2023}, and later adopted as a change detection task in the PANGAEA benchmark \citep{marsocciPANGAEAGlobalInclusive2024}. We compared annual embedding performance on the HLS Burn Scars benchmark with published results from the PANGAEA benchmark and the Prithvi paper. This comparison used the original HLS Burn Scars masks and the original PANGAEA train/validation split, to match the published protocol. We restricted the evaluation to the single-fire subset. We evaluated Tessera and AlphaEarth embeddings on the single-fire subset using logistic regression, RF, and the lightweight U-Net. We reported mean metrics averaged across the burned and unburned classes, matching the benchmark reporting convention. Mean metrics are generally higher because the unburned class is easier to classify and represents the majority of pixels. We compared these results with published PANGAEA baselines and the best-performing geospatial foundation model under the 100\% training data setting. Because the published results were evaluated on the full benchmark rather than our single-fire subset, this comparison was intended as contextual rather than a strict head-to-head benchmark.

\subsection{Results}

Lightweight downstream models trained on Tessera embeddings reached or exceeded published results on the HLS Burn Scars benchmark (Table~\ref{tab:pangaea_comparison}). A Random Forest on Tessera embeddings (mIoU = 0.850, m-F1 = 0.915) performed on par with the supervised U-Net baseline, which remains the strongest published PANGAEA result and outperformed all benchmarked geospatial foundation models on this task. Tessera with the lightweight U-Net (mIoU = 0.922, m-F1 = 0.959) exceeded the PANGAEA U-Net baseline, the best-performing PANGAEA GFM (Prithvi), and the fine-tuned Prithvi result from the original benchmark paper.

\begin{table}[H]
\centering
\caption{Contextual comparison with published PANGAEA HLS Burn Scars benchmark results. Metrics are mean class scores. Published values are reported from the original benchmarks over the full chip set and are not strictly head-to-head with our single-fire subset evaluation; our own post-fire HLS U-Net, evaluated on both sets, quantifies the offset between them.}
\label{tab:pangaea_comparison}
\begin{tabular}{lcc}
\toprule
Approach & mIoU $\uparrow$ & m-F1 $\uparrow$ \\
\midrule
\multicolumn{3}{l}{\textit{Annual embedding approach (single-fire subset)}} \\
Tessera + LogReg & 0.778 & 0.866 \\
Tessera + RF & 0.850 & 0.915 \\
\textbf{Tessera + Light U-Net} & \textbf{0.922} & \textbf{0.959} \\
AlphaEarth + Light U-Net & 0.792 & 0.875 \\
\midrule
\multicolumn{3}{l}{\textit{Our post-fire HLS baseline}} \\
Post-fire HLS + U-Net (full benchmark) & 0.861 & 0.920 \\
Post-fire HLS + U-Net (single-fire subset) & 0.886 & 0.937 \\
\midrule
\multicolumn{3}{l}{\textit{Published post-fire HLS benchmarks (full benchmark)}} \\
PANGAEA U-Net baseline & 0.845 & 0.911 \\
PANGAEA best-performing GFM (Prithvi) & 0.827 & 0.899 \\
Prithvi (original paper, fine-tuned) & 0.848 & 0.914 \\
\bottomrule
\end{tabular}
\end{table}

This performance was achieved with substantially smaller trainable models and simpler inputs. PANGAEA evaluates frozen GFM encoders through a trainable UPerNet decoder, with 30.9--164.4M trainable parameters for GFM configurations and 14.8M for its U-Net baseline. Our lightweight U-Net trains approximately 0.87M parameters and the Random Forest operates on individual pixels without a spatial decoder.

Note, however, that the comparison is contextual rather than strictly head-to-head. Published PANGAEA results cover the full benchmark, whereas our evaluation was restricted to the single-fire subset to avoid label mismatch between event-specific masks and annual embeddings (Section~\ref{sec:delineation-experiment}). To bridge the two, we trained our own post-fire HLS U-Net baseline under both protocols. On the full benchmark it closely reproduced the published PANGAEA U-Net result (m-F1 = 0.920 vs.\ 0.911), validating our implementation; trained and evaluated on the single-fire subset, the same architecture scored modestly higher (m-F1 = 0.937), indicating the subset is somewhat easier. Despite this, the Tessera margin over published results exceeded this offset.
\section{Effect of explicit multi-year temporal context}
\label{supp:multiyear}

Annual embeddings may encode temporal change within a single yearly representation, but disturbance information could also be recovered by explicitly comparing embeddings across years. We therefore tested whether adjacent-year embeddings or explicit embedding differences improved burned area classification relative to the fire-year embedding alone. For Tessera and AlphaEarth, we evaluated six temporal feature configurations: the fire-year embedding $E_0$; the difference between the fire-year and pre-fire embeddings, $E_0-E_{-1}$; the concatenated pre-fire and fire-year embeddings, $[E_{-1},E_0]$; the same pair together with their difference, $[E_{-1},E_0,E_0-E_{-1}]$; a three-year stack, $[E_{-1},E_0,E_{+1}]$; and a pre/post stack excluding the fire year, $[E_{-1},E_{+1}]$.

All configurations were evaluated using Random Forest classifiers with the same hyperparameters as the main benchmark experiments: 200 trees, $\sqrt{d}$ features considered per split, a minimum leaf size of 5, and a minimum split size of 10. We used the same 793-chip cohort and three-fold spatial cross-validation scheme as the annual fire-mapping experiment, with 528--531 training chips and 262--266 validation chips per fold. Models were trained on balanced samples of 1{,}000 burned and 1{,}000 unburned pixels per training chip using the refined MTBS labels. The $E_0$ results were obtained from the corresponding annual fire-mapping models, while a separate classifier was trained for each additional temporal configuration.

To evaluate all configurations using a common and computationally tractable protocol, we uniformly sampled 10{,}000 valid pixels from each chip using a fixed random seed. Evaluation samples were not class-balanced and contained approximately 11.4\% burned pixels. Within each fold, models were evaluated only on samples from the held-out chips. The same sampled locations were used across temporal configurations, and samples with non-finite embedding values were excluded. F1, IoU, and AUROC were calculated by pooling sampled pixels across the validation chips in each fold and are reported as the mean and standard deviation across the three folds.

\subsection{Results}

Explicit temporal context affected Tessera and AlphaEarth differently (Figure~\ref{fig:supp_temporal_context}; Table~\ref{tab:supp_multiyear}). For Tessera, the fire-year embedding alone already contained most of the recoverable burned area signal, achieving an F1 score of $0.840\pm0.011$. Adding the pre-fire embedding and its explicit difference from the fire year increased F1 only modestly to $0.857\pm0.009$. The concatenated pre-fire and fire-year embeddings and the three-year stack performed similarly, with F1 scores of $0.849\pm0.012$ and $0.855\pm0.011$, respectively. Using only the temporal difference did not improve performance ($0.835\pm0.011$), while excluding the fire-year embedding reduced F1 substantially to $0.676\pm0.013$.

AlphaEarth benefited considerably more from explicit temporal context. Adding the pre-fire embedding and the explicit temporal difference increased F1 from $0.583\pm0.041$ for $E_0$ alone to $0.731\pm0.022$. The temporal difference alone and the three-year stack also improved performance, reaching F1 scores of $0.676\pm0.026$ and $0.682\pm0.023$, respectively. When the fire-year embedding was excluded, the pre/post stack performed similarly to $E_0$ alone ($0.599\pm0.025$ versus $0.583\pm0.041$).

IoU followed the same overall pattern. AUROC varied little among Tessera configurations containing $E_0$, ranging from 0.973 to 0.975, whereas AlphaEarth AUROC increased from $0.908\pm0.007$ for $E_0$ alone to $0.949\pm0.004$ with the full temporal context. For the AlphaEarth configurations for which complete-chip results were available, F1 from the uniform evaluation sample differed by only 0.007--0.015, indicating that the sampling protocol closely approximated complete-chip evaluation.

Together, these results suggest that wildfire disturbance information is concentrated within Tessera's fire-year representation, so explicitly supplying adjacent annual states provides little additional benefit. AlphaEarth also retains useful disturbance information, but that information is extracted more effectively when temporal contrast is supplied explicitly to the downstream model.

\begin{figure}[H]
\centering
\includegraphics[width=\textwidth]{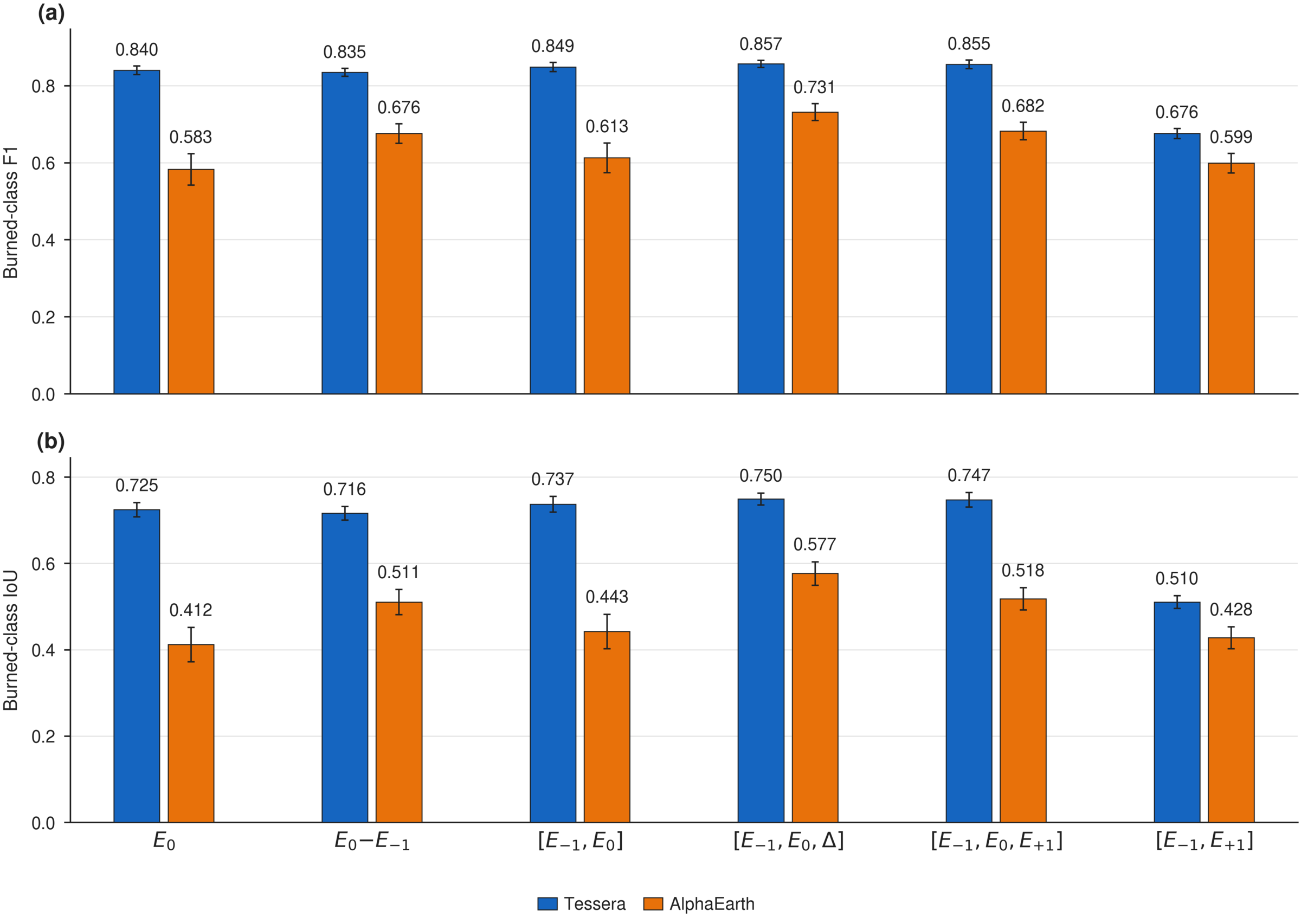}
\caption{Effect of explicit multi-year temporal context on burned area classification using Tessera and AlphaEarth embeddings. Bars show mean (a) burned-class F1 and (b) burned-class IoU across three spatial cross-validation folds; numbers above the bars give the corresponding means, and error bars indicate one standard deviation. Models were trained on balanced burned and unburned pixel samples and evaluated using 10{,}000 uniformly sampled valid pixels from each held-out chip. $E_{-1}$, $E_0$, and $E_{+1}$ denote the pre-fire, fire-year, and post-fire annual embeddings, respectively, and $\Delta=E_0-E_{-1}$.}
\label{fig:supp_temporal_context}
\end{figure}

\begin{table}[H]
\centering
\caption{Effect of explicit multi-year temporal context on burned-area classification using Tessera and AlphaEarth embeddings. Random Forest models were evaluated using 10{,}000 uniformly sampled valid pixels per validation chip under spatially blocked three-fold cross-validation. Values are fold mean $\pm$ standard deviation. $E_0$ denotes the fire-year embedding, $E_{-1}$ the preceding year, $E_{+1}$ the following year, and $\Delta=E_0-E_{-1}$. Bold values indicate the highest F1 and IoU for each embedding product.}
\label{tab:supp_multiyear}
\scriptsize
\setlength{\tabcolsep}{3pt}
\resizebox{\textwidth}{!}{%
\begin{tabular}{@{}lcccccc@{}}
\toprule
Features & Tessera F1 & Tessera IoU & Tessera AUROC & AlphaEarth F1 & AlphaEarth IoU & AlphaEarth AUROC \\
\midrule
$E_0$ & $0.840 \pm 0.011$ & $0.725 \pm 0.016$ & $0.973 \pm 0.002$ & $0.583 \pm 0.041$ & $0.412 \pm 0.040$ & $0.908 \pm 0.007$ \\
$E_0-E_{-1}$ & $0.835 \pm 0.011$ & $0.716 \pm 0.016$ & $0.963 \pm 0.002$ & $0.676 \pm 0.026$ & $0.511 \pm 0.029$ & $0.930 \pm 0.004$ \\
$[E_{-1},E_0]$ & $0.849 \pm 0.012$ & $0.737 \pm 0.018$ & $0.973 \pm 0.002$ & $0.613 \pm 0.039$ & $0.443 \pm 0.040$ & $0.915 \pm 0.006$ \\
$[E_{-1},E_0,\Delta]$ & $\mathbf{0.857 \pm 0.009}$ & $\mathbf{0.750 \pm 0.014}$ & $0.974 \pm 0.002$ & $\mathbf{0.731 \pm 0.022}$ & $\mathbf{0.577 \pm 0.027}$ & $0.949 \pm 0.004$ \\
$[E_{-1},E_0,E_{+1}]$ & $0.855 \pm 0.011$ & $0.747 \pm 0.017$ & $0.975 \pm 0.002$ & $0.682 \pm 0.023$ & $0.518 \pm 0.026$ & $0.944 \pm 0.002$ \\
$[E_{-1},E_{+1}]$ & $0.676 \pm 0.013$ & $0.510 \pm 0.015$ & $0.940 \pm 0.003$ & $0.599 \pm 0.025$ & $0.428 \pm 0.025$ & $0.921 \pm 0.001$ \\
\bottomrule
\end{tabular}%
}
\end{table}
\section{False-negative rates by burn severity}
\label{supp:severity}
\label{supp:severity-fnr}

\begin{table}[H]
\centering
\caption{False-negative rates by MTBS burn-severity class for Tessera and AlphaEarth using the Lightweight U-Net. Values are aggregated across all 793 benchmark chips using out-of-fold predictions from the annual fire mapping experiment (Section~\ref{sec:annual-fire-mapping-results}). False-negative rates (FNR) are calculated over burned pixels within each severity class; the final column reports the AlphaEarth--Tessera difference in percentage points.}
\label{tab:supp_severity_fnr}
\begin{tabular}{@{}lrrrr@{}}
\toprule
Severity & $n$ burned & Tessera FNR & AlphaEarth FNR & Gap (AE $-$ Tessera) \\
\midrule
Low      & 16.8\,M & 10.8\% & 29.0\% & $+18.2$\,pp \\
Moderate & 4.9\,M  & 1.3\%  & 6.3\%  & $+5.1$\,pp \\
High     & 1.6\,M  & 0.1\%  & 0.8\%  & $+0.6$\,pp \\
\bottomrule
\end{tabular}
\end{table}

\begin{table}[H]
\centering
\caption{False-negative rates by MTBS burn-severity class for Tessera and AlphaEarth using logistic regression. Values are aggregated across all 793 benchmark chips using out-of-fold predictions from the annual fire mapping experiment (Section~\ref{sec:annual-fire-mapping-results}). False-negative rates (FNR) are calculated over burned pixels within each severity class; the final column reports the AlphaEarth--Tessera difference in percentage points.}
\label{tab:supp_severity_fnr_logreg}
\begin{tabular}{@{}lrrr@{}}
\toprule
Severity & Tessera FNR & AlphaEarth FNR & Gap (AE $-$ Tessera) \\
\midrule
Low      & 12.2\% & 28.3\% & $+16.1$\,pp \\
Moderate & 3.8\%  & 9.5\%  & $+5.7$\,pp \\
High     & 2.1\%  & 3.7\%  & $+1.6$\,pp \\
\bottomrule
\end{tabular}
\end{table}
\section{Overview of supplementary-only comparisons and ablations}
\label{supp:cohorts}

\begin{table}[H]
\centering
\caption{Overview of supplementary-only comparisons and ablations. Complete protocols are provided in the linked sections.}
\label{tab:cohorts_supp}
\small
\renewcommand{\arraystretch}{1.15}
\begin{tabularx}{\textwidth}{
    >{\raggedright\arraybackslash}p{0.34\textwidth}
    >{\raggedright\arraybackslash}X}
\toprule
\textbf{Analysis} & \textbf{Cohort and evaluation} \\
\midrule
\textbf{Published-transfer comparison (Section~\ref{supp:seydi})} &
Same US-trained Tessera model evaluated without retraining on 17 EMSR fires from the published AlphaEarth comparison; 15 belong to the main cohort and two are additional 2022 fires \\
\addlinespace[4pt]
\textbf{PANGAEA comparison (Section~\ref{supp:pangaea})} &
Single-fire subset: 588 chips; official 394/194 split; original HLS Burn Scars masks; full-chip, mean-class scoring \\
\addlinespace[4pt]
\textbf{Multi-year context ablation (Section~\ref{supp:multiyear})} &
793 chips; six temporal configurations; same three-fold spatial cross-validation as annual fire mapping; RF trained on balanced samples and evaluated using 10{,}000 uniformly sampled valid pixels per validation chip \\
\bottomrule
\end{tabularx}
\end{table}

\bibliographystyle{elsarticle-harv}
\bibliography{references,manual}
\fi

\end{document}